\documentclass[runningheads]{llncs}

\usepackage{graphicx}
\usepackage{times}
\usepackage{helvet}
\usepackage{courier}
\usepackage[utf8]{inputenc}
\usepackage{wrapfig}
\usepackage{fancyhdr}
\usepackage{algorithm}
\usepackage{amssymb}
\usepackage{listings}
\usepackage{todonotes}
\usepackage{mathtools}
\usepackage{stmaryrd}
\usepackage{mathptmx}
\usepackage{hyperref}
\usepackage{comment}
\usepackage{subcaption}
\usepackage{centernot}
\usepackage{balance}
\usepackage{eso-pic}
\usepackage{tikz}
\usepackage{multirow}
\usepackage{xspace}
\usepackage{nicefrac}
\usepackage[normalem]{ulem}
\usepackage[title, titletoc, page]{appendix}
\usetikzlibrary{calc,arrows,shapes,snakes,automata,backgrounds,petri}
\usepackage[nameinlink]{cleveref}
\usepackage{thmtools,thm-restate}
\usepackage{adjustbox}
\usepackage{url}
\usepackage{breakurl}
\usepackage{algpseudocode}
\usepackage{chngcntr}
\usepackage{tabularx}
\usepackage{yfonts}
\usepackage{mathrsfs}

\usetikzlibrary{positioning}
\usepackage[dvipsnames, table]{xcolor}
\definecolor{dtu_red_1}{RGB}{153, 0, 0}
\definecolor{dtu_red_2}{RGB}{222, 203, 203}
\definecolor{dtu_red_3}{RGB}{240, 231, 231}
\usepackage{pgfplots}
\usepgfplotslibrary{external}

\newcommand{\ie}{i.e.\xspace}
\newcommand{\eg}{e.g.\xspace}

\newcommand{\Dist}[1]{\mathit{Dist}(#1)}
\newcommand{\Strat}{\mathfrak{S}}
\newcommand{\OStrat}{\mathfrak{O}}
\newcommand{\Act}{\mathit{Act}}
\newcommand{\rew}{\mathit{rew}}
\newcommand{\Paths}[1]{\mathit{Paths}(#1)}
\newcommand{\OPaths}[1]{\mathit{OPaths}(#1)}
\newcommand{\Rew}[2]{\mathit{rew}_{#1}(#2)}
\newcommand{\Prob}[3]{P_{#1}^{#2}(#3)}

\newcommand{\ExpRew}[2]{\textsf{ExpRew}^{#2}(#1)}
\newcommand{\MinExpRew}[1]{\textsf{MinExpRew}(#1)}

\newcommand{\StratP}{\mathfrak{S}_{p}}
\newcommand{\MinExpRewP}[1]{\textsf{MinExpRew}_{p}(#1)}

\newcommand{\StratPD}{\mathfrak{S}_{pd}}
\newcommand{\MinExpRewPD}[1]{\textsf{MinExpRew}_{pd}(#1)}

\newcommand{\OStratP}{\mathfrak{O}_{p}}
\newcommand{\OStratPD}{\mathfrak{O}_{pd}}

\newcommand{\mdp}{M}
\newcommand{\pomdp}{\mathcal{M}}
\newcommand{\pmc}{\mathcal{D}}

\newcommand{\obs}{\mathit{obs}}

\newcommand{\Poly}[1]{\mathbb{Q}[#1]}
\newcommand{\instance}[2]{#1[#2]}

\newcommand{\goalColor}{green!60!black}
\newcommand{\obsGoal}{{\color{\goalColor}\checkmark}}

\newcommand{\setObs}[1]{\langle{#1}\rangle}
\newcommand{\OO}{O_\obsGoal}
\newcommand{\OOp}{O'_\obsGoal}

\newcommand{\mdpline}{\mdp_{\text{line}}}

\newcommand{\Line}{$\mathsf{Line}$\xspace}
\newcommand{\Grid}{$\mathsf{Grid}$\xspace}
\newcommand{\Maze}{$\mathsf{Maze}$\xspace}

\newcommand{\POP}{\textbf{POP}\xspace}
\newcommand{\SSP}{\textbf{SSP}\xspace}
\newcommand{\OOP}{\textbf{OOP}\xspace}

\newcommand{\zthree}{\textsc{z3}\xspace}
\newcommand{\tpmc}{tpMC\xspace}

\newcommand{\mdptrap}{\mdp_{\text{trap}}}

\newcommand{\obsGoalSS}{{\color{\goalColor}\scriptsize\checkmark}}

\renewcommand{\subsubsection}[1]{\smallskip\noindent\textbf{#1}}
\newcolumntype{C}[1]{>{\centering\arraybackslash}m{#1}}

\AddToHook{cmd/appendix/before}{\crefalias{section}{appendix}}
\newcommand{\appref}[1]{\hyperref[#1]{Appendix~\ref*{#1}}}

\begin{document}
%

\title{Scaling Observation-aware Planning in Uncertain Domains \thanks{This work has been partially supported by the Danish Data Science Academy (DDSA) through travel grants to the conference AISoLA 2025 in Rhodes, Greece.}}
%
%

\author{
Adrian Zvizdenco \and
Arthur Conrado Veiga Bosquetti \and \\
Alberto Lluch Lafuente\orcidID{0000-0001-7405-0818} \and \\
Christoph Matheja\orcidID{0000-0001-9151-0441}
}
\authorrunning{Adrian Zvizdenco, Arthur Bosquetti, Alberto Lluch Lafuente, et al.}
%
\institute{
  Technical University of Denmark, Kongens Lyngby\\
 \email{\{s204683, s204718, albl, chmat\}@dtu.dk} 
}

\maketitle              
%

\begin{abstract}

Deciding which sensing capabilities to deploy on an agent in uncertain domains is a fundamental engineering challenge, in which one balances task achievability against the high costs of hardware and processing. This problem has previously been formalized as the Optimal Observability Problem (OOP) \cite{cav-base-oop}, based on the well-known Partially Observable Markov Decision Process (POMDP) model for decision-making.
This work studies (sub-)symbolic techniques to scale solving of decidable fragments of the \OOP, namely the Sensor Selection Problem (SSP) and the Positional Observability Problem (POP). 
Besides improving the original approach based on parameter synthesis, we develop a new solving method that identifies sensible observation functions via decomposition of POMDPs, improving performance by 3 and 5 orders of magnitude for instance size and runtime, respectively.
\end{abstract}
\section{Introduction}\label{sec:introduction}

Decision-making agents rely on their sensing capabilities to navigate the surrounding world and reach their designated goals. However, information is rarely free or complete.
A critical trade-off in domains such as security \cite{alyzia-phd}, cyber-physical systems \cite{DBLP:conf/staf/JdeedSBSPCSBPE19}, and planning for autonomous systems \cite{zilberstein-96} is thus:
how can one design a system's sensing capabilities to ensure a task is achievable without overspending?


Konsta et al.~\cite{cav-base-oop} formalised this question as the \textit{Optimal Observability Problem} (OOP) for Partially Observable Markov Decision Processes (POMDP).
While traditional research on POMDPs focuses on how an agent should act given a concrete set of observations, 
the  \OOP considers the system design phase and asks how to assign observations (think: sensors) cost-effectively. While  \OOP is undecidable in general, there are decidable fragments for restricted classes of strategies, particularly 
the Sensor Selection (SSP) and the Positional Observability (POP) problem (cf.~\cite{cav-base-oop}).

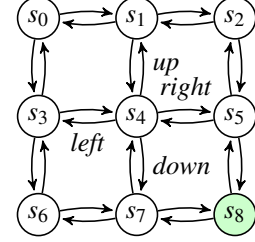
\begin{wrapfigure}{r}{0.28\textwidth}
\vspace{-0.84cm}
\begin{tikzpicture}[->,>=stealth',shorten >=1pt,auto,node distance=1.3cm,
                    semithick]
  \tikzstyle{every state}=[fill=white,draw=black,text=black]

  \node[state,inner sep=2pt,minimum size=1pt]         (S0)                   {$s_0$};
  \node[state,inner sep=2pt,minimum size=1pt]         (S1) [below of=S0]     {$s_3$};
  \node[state,inner sep=2pt,minimum size=1pt]         (S2) [below of=S1]     {$s_6$};
  \node[state,inner sep=2pt,minimum size=1pt]         (S3) [right of=S0]     {$s_1$};
  \node[state,inner sep=2pt,minimum size=1pt]         (S4) [right of=S1]     {$s_4$};
  \node[state,inner sep=2pt,minimum size=1pt]         (S5) [right of=S2]     {$s_7$};
  \node[state,inner sep=2pt,minimum size=1pt]         (S6) [right of=S3]     {$s_2$};
  \node[state,inner sep=2pt,minimum size=1pt]         (S7) [right of=S4]     {$s_5$};
  \node[state,inner sep=2pt,minimum size=1pt][fill=green!20]  (S8) [right of=S5]     {$s_8$};

  \path (S0) edge [bend left=10]             node {} (S3)
             edge  [bend right=10]             node [left]{} (S1)
        (S1) edge   [bend left=10]            node {} (S4)
             edge  [bend right=10]             node [above left] {} (S2)
             edge  [bend right=10]             node [right] {} (S0)
        (S2) edge     [bend left=10]          node {} (S5)
             edge    [bend right=10]           node [below right]{} (S1)
        (S3) edge    [bend right=10]           node [above left]{} (S4)
             edge   [bend left=10]            node {} (S6)
             edge     [bend left=10]          node {} (S0)
        (S4) edge    [bend left=10]           node {$\mathit{right}$} (S7)
             edge   [bend left=10]            node {$\mathit{down}$} (S5)
             edge   [bend left=10]            node {$\mathit{left}$} (S1)
             edge    [bend right=10]           node [ right] {$\mathit{up}$} (S3)
        (S5) edge   [bend left=10]            node {} (S4)
        edge   [bend left=10]            node {} (S8)
             edge     [bend left=10]          node {} (S2)
        (S6) edge    [bend left=10]           node {} (S7)
             edge    [bend left=10]           node {} (S3)
        (S7) edge     [bend left=10]          node {} (S4)
             edge     [bend left=10]          node {} (S8)
             edge    [bend left=10]           node {} (S6)
        (S8) edge    [bend left=10]           node  [above left]{} (S7)
             edge    [bend left=10]           node {} (S5);
\end{tikzpicture}
\caption{3x3 grid world.} 
\vspace{-0.7cm}
    \label{grid1}
\end{wrapfigure}

As an illustration, consider sensor selection in the grid world ~\cite{LITTMAN1995362} in \Cref{grid1}. An agent is placed randomly on one of the non-goal states ($s_0$-$s_7$), with the objective of reaching the goal $s_8$ (marked by the green color). The agent can navigate the world using the actions $\mathit{left}, \mathit{right}, \mathit{up}, \mathit{down}$, available in each state. Every time the agent picks an action, it takes
one step. 
In \SSP, the agent determines its exact location only if it observes a state with an activated sensor. Otherwise, the agent can only guess a belief distribution over the states with their sensor turned off.
With an unlimited budget, one achieves full observability by placing a sensor in each world position. In this case, the problem becomes trivial: The agent reaches the goal in an (optimal) expected number of $2.25$ steps. 
For a restricted budget, we shall identify which locations provide the most meaningful information to guide the agent. In this case, $2$ sensors (e.g. on $s_2$, $s_5$) suffice such that the agent still achieves the optimal expected number of steps.


Konsta et al. \cite{cav-base-oop} developed decision procedures for fragments of the  \OOP problem by reducing it to parameter synthesis for Markov chains (cf.~\cite{junges2020parameter}), implemented via  probabilistic model checking (PMC) or SMT solving.
Our aim is to leverage symbolic and sub-symbolic AI techniques to solve  \OOP problems at scale.

\subsubsection{Contributions.}
We study alternative SMT encodings and relaxation techniques for the  \OOP solving approach from \cite{cav-base-oop} based on parameter synthesis (\Cref{sec:enhancements-smt}).
Our implementation can solve  \OOP instances from \cite{cav-base-oop} $\sim 10^3$ times faster. It also scales to larger instances by a factor of $\sim 75$. 
\Cref{sec:integration-pomdp-oracles} considers symbolic oracles for the reward of instantiated POMDPs in the optimal observability context. 
This results in Heuristic-based Enumeration for  \OOP instances via POMDP decomposition (\Cref{sec:informed-search-algorithms}).
%
The resulting algorithm improves upon our own enhancements in \Cref{sec:enhancements-smt} by factors of $10^3$ and $10^2$ for solving time and instance size, respectively.
To further reduce the search space,  \Cref{sec:informed-search-algorithms} explores a procedure for initialising observation functions.


\subsubsection{Related Work.}
This work builds a continuation of the prior research and \OOP formalisations from \cite{cav-base-oop}.
As in the base paper, we ought to mention the sensor synthesis problem based on reachability objectives introduced in \cite{chatterjee2018sensor}. Important distinctions in their work refer to partially initialised observation functions and a focus on \textit{qualitative} POMDP objectives, rather than \textit{quantitative} objectives, \ie the minimal expected reward we consider. Similarly, the choice of SMT encodings over the SAT-based solution in \cite{chatterjee2018sensor} reflects an increase in the complexity class one is operating with in this work (from \textsc{np}-complete to \textsc{etr}-complete).
The survey by Junges et al. \cite{jansen2022-parameter-synthesis-survey} provides a comprehensive overview of the techniques used to analyse parametric models, involving formulations of various parameter synthesis problems (feasibility, region verification and partitioning). They discuss the trade-offs between symbolic PMC approaches and sampling-based methods, and establish a state-of-the-art procedure for feasibility checking. We deem it possible to convert \OOP statements as a query for Region Verification, but the method is only effective for a few parameters, as concluded in \cite{jansen2022-parameter-synthesis-survey}.
Since it does not scale for our setting, we aim to investigate other sampling approaches for optimising parameters. Finally, the survey relates the POMDP controller synthesis to the feasibility problem in the parametric MC, and discusses complexity results depending on the considered memory bounds. This analysis is  foundational to our work.
Recent research on (PO)MDP policy synthesis~\cite{robust-smt-pmc-2025} aims to fulfill structural constraints and robustness against model perturbations. 
The SAT-Modulo-PMC framework~\cite{robust-smt-pmc-2025} integrates combinatorial SMT solving with precise reward metrics, which are evaluated in the Storm probabilistic model checker \cite{storm} and injected into a new dedicated \zthree theory. The approach significantly outperforms traditional synthesis methods. However, the optimal observability problem considered in our work is fundamentally different. 

\section{Preliminaries}
\vspace{-0.1cm}
\label{sec:preliminaries}

We briefly recap the formalisation of the \OOP problem and its decidable fragments that we aim to solve more efficiently. 
A detailed treatment is found in \cite{cav-base-oop}.


\subsection{Partially Observable Markov Decision Processes}
\label{POMDPs}

A partially observable Markov decision process (POMDP) is an MDP with imperfect information about the current state, that is, certain states are indistinguishable. The standard model is extended with rewards and dedicated goal states.

\begin{definition}[POMDPs]\label{def:POMDP}
A \emph{partially observable Markov decision process} is a tuple 
$\pomdp = (M, O, \obs)$, where the tuple $M = (S, I, G, \Act, P, \rew)$ is an MDP, $O$ is a finite set of \emph{observations}, and $\obs\colon S \to O \uplus \{ \obsGoal \}$ is an \emph{observation function} such that \mbox{$\obs(s) = \obsGoal$ iff $s \in G$.}\footnote{Here, $A \uplus B$ denotes the union $A \cup B$ of sets $A$ and $B$ if $A \cap B = \emptyset$; otherwise, it is undefined.} Hereby, $S$ is a finite set of \emph{states}, $I\subseteq S$ is a set of (uniformly distributed) \emph{initial states}, $G\subseteq S$ is a set of \emph{goal states}, $Act$ is a finite set of \emph{actions}, $P\colon S\times Act \rightarrow Dist(s)$ is a \emph{transition probability function}, and $rew\colon S\rightarrow \mathbb{R}_{\geq 0}$ is a \emph{reward function}.
\end{definition}

For simplicity, we use a dedicated observation $\obsGoal$ for goal states and only consider observation functions of the above kind.
We write $\OO$ as a shortcut for $O \uplus \{\obsGoal\}$ and refer to components of an MDP $M$ by $S_M$, $P_M$, and so on.

%
\begin{figure}[ht]{}
\vspace{-0.4cm}
\centering
\begin{tikzpicture}[->,>=stealth',shorten >=1pt,auto,node distance=1.8cm, semithick]

  \tikzstyle{every state}=[fill=white,draw=black,text=black]

  \node[state,inner sep=2pt,minimum size=1pt, fill=blue!20]         (S0)                   {$s_0$};
  \node[state,inner sep=2pt,minimum size=1pt, fill=red!20]         (S1) [right of=S0]     {$s_1$};
  \node[state,inner sep=2pt,minimum size=1pt][fill=green!20]          (S2) [right of=S1]     {$s_2$};
  \node[state,inner sep=2pt,minimum size=1pt, fill=red!20]         (S3) [right of=S2]     {$s_3$};
  \node[state,inner sep=2pt,minimum size=1pt, fill=blue!20]         (S4) [right of=S3]     {$s_4$};

  \node[below=0.15cm of S0] (s1) {$@ s_0$};
  \node[below=0.15cm of S1] (s2) {$@ s_1$};
  \node[below=0.15cm of S3] (s3) {$@ s_3$};
  \node[below=0.15cm of S4] (s4) {$@ s_4$};

  \path (S0) edge   [bend left=20]           node [above]{$r$ : $p$} (S1)
             edge   [loop above]             node {$r$ : $1-p$} (S0)
             edge   [loop left]              node [left] {$\ell$ : 1\, } (S0)
        (S1) edge                            node {$r$ : $p$} (S2)
             edge   [bend left=20]           node {$\ell$ : $p$} (S0)
             edge   [loop above]             node {$\ell$,$r$ : $1-p$} (S1)
        (S2) edge   [loop above]             node [above]{$\ell$,$r$ : 1} (S2)
        (S3) edge   [bend left=20]           node {$r$ : $p$} (S4)
             edge                            node [above]{$\ell$ : $p$} (S2)
             edge   [loop above]             node {$\ell$,$r$ : $1-p$} (S3)
        (S4) edge   [bend left=20]           node {$\ell$ : $p$} (S3)
             edge   [loop above]             node {$\ell$ : $1-p$} (S4)
             edge   [loop right]             node [right]{\, $r$ : 1} (S4);
\end{tikzpicture}
\vspace{-0.1cm}
\caption{POMDP $\pomdp_{line}$ with $\obs=\{\{s_0, s_4\}\mapsto{\color{blue!70!black}o_1},\{s_1, s_3\}\mapsto{\color{red!70!black}o_2}, \{s_2\}\mapsto \obsGoal \}$  and underlying MDP $\mdpline$ for some fixed constant $p \in [0,1]$; the initial states are $s_0,s_1,s_3,s_4$. The @ labels identify the observable locations in the instance.}
\label{POMDPprobs}
\vspace{-0.7cm}
\end{figure}
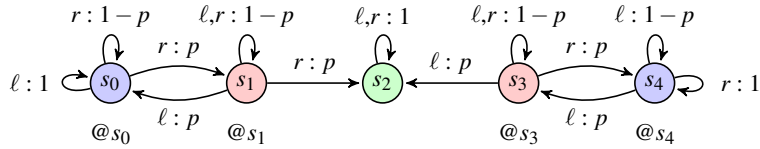
\begin{example}\label{ex:2-pomdp}
Consider a line example involving an agent that is placed at one of the four random locations. The agent needs to reach a {\color{\goalColor} goal} by moving to the $\ell$(eft) or $r$(ight). Whenever it decides to move, it successfully does so with some fixed probability $p \in [0,1]$; with probability $1-p$, it stays at the current location due to a failure.
  \Cref{POMDPprobs} depicts a POMDP $\pomdp_{line}$ modeling the above scenario using five states $s_0$-$s_4$. Here, ${\color{\goalColor}s_2}$ is the single goal state. All other states are initial. An edge $s_i \xrightarrow{\alpha: q} s_j$ indicates that $P(s_i, \alpha)(s_j) = q$. 
  The reward (omitted in \Cref{POMDPprobs}) is $0$ for ${\color{\goalColor}s_2}$, and $1$ for all other states.
The colors in \Cref{POMDPprobs} indicate a POMDP obtained from $\mdpline$ by assigning observations {\color{blue!70!black}$o_1$} to $s_0$ and $s_4$, {\color{red!70!black}$o_2$} to $s_1$ and $s_4$, and $\obsGoal$ to {\color{\goalColor}$s_2$}.
Hence, we know how far away from the goal state {\color{\goalColor}$s_2$} we are but not which action leads to the goal.
An optimal, positional, and deterministic strategy $\sigma$ for the MDP $\mdpline$ (\Cref{POMDPprobs}) chooses action $r$(ight) for states $s_0$, $s_1$ and $\ell$(eft) for $s_3$, $s_4$. 
  For $p = \nicefrac{2}{3}$, the (minimal) expected number of steps until reaching {\color{\goalColor}$s_2$} is $\ExpRew{\mdpline}{\sigma} = 3$.
\end{example}

In a POMDP $\pomdp$, we assume that we cannot directly see a state, say $s$, but only its assigned observation $\obs_\pomdp(s)$ -- all states in $\obs^{-1}_{\pomdp}(o) = \{ s ~|~ \obs_\pomdp(s) = o \}$ thus become indistinguishable.
Consequently, multiple path fragments in the underlying MDP $M$ might also become indistinguishable.
More formally, the \emph{observation path fragment} $\obs_\pomdp(\pi)$ of a path fragment $\pi = s_0 \alpha_0 s_1 \ldots s_n \in \Paths{M}$ is defined as
\[
\obs_\pomdp(\pi) = \obs_\pomdp(s_0)\, \alpha_0\, \obs_\pomdp(s_1)\, \ldots\, \obs_\pomdp(s_n).
\]
We denote by $\OPaths{\pomdp}$ the set of all observation paths obtained from the paths of $\pomdp$'s underlying MDP $M$, i.e. $\OPaths{\pomdp} = \{ \obs_\pomdp(\pi) ~|~ \pi \in \Paths{M} \}$.

\emph{Strategies} for POMDPs are defined as for their underlying MDPs, which map non-empty finite sequences of states to distributions over actions $\sigma\colon S_M^{+} \to Dist(\Act_M)$. We denote by $\mathfrak{S}(M)$ the set of all strategies for MDP $M$. However, POMDP strategies must be \emph{observation-based}, that is, they have to make the same decisions for path fragments that have the same observation path fragment. We denote by $\OStrat(\pomdp)$ the set of all observation-based strategies of $\pomdp$.

\begin{definition}[Observation-Based Strategies]
    An \emph{observation-based strategy} $\sigma$ for a POMDP $\pomdp = (M, O, \obs)$ is a function $\sigma \colon \OPaths{\pomdp} \to \Dist{\Act_M}$ such that:
    \vspace{-0.1cm}
    \begin{itemize}
        \item $\sigma$ is a strategy for the MDP $M$, i.e. $\sigma \in \Strat(M)$ and
        \item for all path fragments $\pi = s_0 \alpha_0 s_1 ... s_n$ and $\pi' = s_0' \alpha_0' s_1' ... s_n'$, if $\obs(\pi) = \obs(\pi')$, then $\sigma(s_0 s_1 \ldots s_n) = \sigma(s_0' s_1' \ldots s_n')$.
    \end{itemize}
\end{definition}

Note that a \emph{path fragment} of an MDP $M$ is a finite sequence $
\pi = s_0\, \alpha_0\, s_1\, \alpha_1\, \ldots\, \alpha_{n-1}\, s_n
$ for some natural number $n$ such that every transition from one state to the next one can be taken for the given action with non-zero probability, i.e. $P_M(s_i,\alpha_i)(s_{i+1}) > 0$ holds for all $i \in \{0, \ldots, n\}$. We denote by $\mathit{first}(\pi) = s_0$ (resp. $\mathit{last}(\pi) = s_n$) the first (resp. last) state in $\pi$. Moreover, we call $\pi$ a \emph{path} if $s_0$ is an initial state, \ie $s_0 \in I_M$, and $s_n \in G_\mdp$ is the first encountered goal state, \ie $s_1, \ldots, s_{n-1} \in S_\mdp \setminus G_M$ and $s_n \in G_\mdp$. We denote by $\Paths{\mdp}$ the set of all paths of $M$, and by $\Paths{\mdp,s}$ - the set of all paths of $M$ originating in $s$. The \emph{cumulative reward} of a path fragment $\pi = s_0\, \alpha_0\, \ldots\, \alpha_{n-1}\, s_n$ of $\mdp$ is the sum of all rewards along the path, that is,
\[
  \Rew{M}{\pi} ~=~ \sum_{i=0}^{n} rew_M(s_i) + \sum_{i=0}^{n}rew_M(s_i,\alpha_i).
\]
Furthermore, for a given strategy $\sigma$, the \emph{probability} of the above path fragment $\pi$ is\footnote{If $\pi$ is a path, notice that our definition does \emph{not} include the probability of starting in $\mathit{first}(\pi)$.}
\[
  \Prob{M}{\sigma}{\pi} ~=~ \prod_{i=0}^{n-1} P_M(s_i, \alpha_i)( s_{i+1}) \cdot \sigma(s_0 \ldots s_i)(\alpha_i).
\]

Put together, the expected reward of $M$ for strategy $\sigma$ is the sum of rewards of all paths weighted by their probabilities, including the initial uniform probability -- at least as long as the goal states are reachable from the initial states with probability one; otherwise, the expected reward is infinite (cf.~\cite{baier2008principles}). Formally, $\ExpRew{M}{\sigma} = \infty$ if $\frac{1}{|I_M|} \cdot \sum_{\pi \in \Paths{M}} \Prob{M}{\sigma}{\pi} < 1$. Otherwise, 
\[
  \ExpRew{M}{\sigma} ~=~ \frac{1}{|I_M|} \cdot \sum_{\pi \in \Paths{M}} \Prob{M}{\sigma}{\pi} \cdot \Rew{M}{\pi}. 
\]
Given an observation-based strategy $\sigma$, we compute the expected reward of a POMDP $\pomdp$ in the same way: $\ExpRew{\pomdp}{\sigma} = \ExpRew{\mdp}{\sigma}$. Similarly, the \textit{value function} of a state $s$ is defined as the expected reward over all paths starting in state $s$ and achieving a goal in $G_\mdp$, and is denoted by $v_s$.
\[
v^{\sigma}_s = \mathbb{E}_{\sigma}\left[ \Rew{M}{\pi} \mid \mathit{first}(\pi) = s\right] = \sum_{\pi \in \Paths{M,s}} \Prob{M}{\sigma}{\pi} \cdot \Rew{M}{\pi}
\]

\begin{definition}[Minimal Expected Reward Restricted to $\chi$]\label{def:min-exp-rew}
The \emph{minimal expected reward} of a POMDP $\pomdp = (M, O, \obs)$ is defined as the infimum among the expected rewards for all observation-based strategies in a set $\chi$:
\[
  \MinExpRew{\pomdp} ~=~ \inf_{\sigma \in \chi} \ExpRew{\pomdp}{\sigma}.
\]
Unless otherwise specified, $\chi$ is assigned to $\OStrat(\pomdp)$. We remind that $\MinExpRew{M}$ of an MDP $M$ is computed analogously, where $\sigma \in \Strat(\mdp)$.
\end{definition}

\noindent\textit{Optimal, positional and deterministic strategies.}
Optimal, positional, and deterministic observation-based strategies for POMDPs are defined analogously to their counterparts for MDPs. In general, strategies may choose actions randomly and based on the history of previously encountered states. We will frequently consider three subsets of strategies.
First, a strategy $\sigma$ for $M$ is \emph{optimal} if it yields the minimal expected reward, \ie $\ExpRew{\pomdp}{\sigma} = \MinExpRew{\pomdp}$. 
Second, a strategy is \emph{positional} if actions depend on the current state only, \ie $\sigma(ws) = \sigma(s)$ for all $w \in S_\pomdp^*$ and $s \in S_\pomdp$.
Third, a strategy is \emph{deterministic} if the strategy always chooses exactly one action, \ie for all $w \in S_\pomdp^+$ there is an $a \in \Act_\pomdp$ such that $\sigma(w) = \delta_a$, where the Dirac distribution $\delta_a$ assigns probability $1$ to the selected action $a$ and probability $0$ to all other actions.
Furthermore, given a positional strategy $\sigma$, we denote by $\pomdp[\sigma] = (M_\pomdp[\sigma], O_\pomdp, \obs_\pomdp)$ the POMDP in which the underlying MDP $M$ is changed to the Markov chain induced by $M$ and $\sigma$.

  When computing expected rewards, we can view a POMDP as an MDP whose strategies are restricted to observation-based ones. Hence, the minimal expected reward of a POMDP is always greater than or equal to the minimal expected reward of its underlying MDP.
  In particular, if there is one observation-based strategy that is also optimal for the MDP, then the POMDP and the MDP have the same minimal expected reward. For MDPs, there always exists an optimal strategy that is also positional and deterministic (cf.~\cite{baier2008principles}). Hence, the minimal expected reward of such an MDP $\mdp$ can alternatively be defined in terms of the expected rewards of its induced Markov chains.

\begin{example}[cntd.]\label{ex:pomdp-strat}
  Consider the POMDP $\pomdp$ in \Cref{POMDPprobs} for $p=\nicefrac{1}{2}$. 
  For the underlying MDP $\mdpline$, we have $\MinExpRew{\mdpline} = 4$.
  Since we cannot reach {\color{\goalColor}$s_2$} from {\color{red!70!black}$s_1$} and {\color{red!70!black}$s_3$} by choosing the same action, every positional and deterministic observation-based strategy $\sigma$ yields $\ExpRew{\pomdp}{\sigma} = \infty$.
  An observation-based positional strategy $\sigma'$ can choose each action with probability $\nicefrac{1}{2}$, which yields $\ExpRew{\pomdp}{\sigma'} = 10$.
  Moreover, for deterministic, but not necessarily positional, strategies,  $\MinExpRew{\pomdp} \approx 4.74$\footnote{Approximate solution provided by \textsc{prism}'s POMDP solver.}.
\end{example}

\smallskip
\noindent\textit{Notation for (PO)MDPs.}
Given a POMDP $\pomdp = (M, O, \obs)$ and an observation function $\obs'\colon S_M \to \OOp$, we denote by $\pomdp\setObs{\obs'}$ the POMDP obtained from $\pomdp$ by setting the observation function to $\obs'$, i.e. $\pomdp\setObs{\obs'} = (M, O', \obs')$.
We call $\pomdp$ 
\emph{fully observable} if all states can be distinguished from one another, i.e. $s_1 \neq s_2$ implies $\obs(s_1) \neq \obs(s_2)$ for all $s_1,s_2 \in S_M$. Throughout this paper, we do not distinguish between a fully-observable POMDP $\pomdp$ and its underlying MDP $M$. 
Hence, we use notation introduced for POMDPs, such as $\pomdp\setObs{\obs'}$, also for MDPs.

\subsection{Optimal Observability Problem \cite{cav-base-oop}}\label{oop}

Konsta et al. \cite{cav-base-oop} discuss observability problems of the form \emph{``what should be observable for a POMDP such that a property of interest can still be guaranteed?''}.
In particular, the \emph{optimal observability problem} is concerned with turning an MDP $\mdp$ into a POMDP $\pomdp = (\mdp,O,\obs)$ such that $\pomdp$'s minimal expected reward remains below a given threshold and, at the same time, the number of available observations $|O|$, i.e. how many non-goal states can be distinguished with certainty, is limited by a budget.
Since every MDP is also a POMDP, this problem has a trivial solution when the observation budget is enough to distinguish non-goal states: We can introduce one observation for every non-goal state, i.e. $O = (S_\mdp \setminus G_\mdp)$, and encode full observability, i.e. $\obs(s) = s$ if $s \in S_\mdp \setminus G_\mdp$ and $\obs(s) = \obsGoal$ if $s \in G_\mdp$.
This decision problem becomes significantly more involved if adding objectives or restricting the admissible observation functions $\obs$, and it is formalised in \cite{cav-base-oop}:
\begin{sloppypar}
\begin{definition}[Optimal Observability Problem (OOP)]\label{def:oo-problem}
  Given a tuple $(\mdp, B, \tau$), where $\mdp$ is an MDP, $B \in \mathbb{N}_{\geq 1}$ is a budget, and $\tau \in \mathbb{Q}_{\geq 0}$ is a (rational) threshold, is there an observation function $\obs\colon S_\mdp \to \OO$ with $|O|\leq B$ such that $\MinExpRew{\mdp\setObs{\obs}} \leq \tau$?

\end{definition}
\end{sloppypar}

\begin{theorem}[Undecidability \cite{cav-base-oop}]
  The optimal observability problem is undecidable.
  \label{th:undecidability}
\end{theorem}

\subsection{Optimal Observability for Positional and Deterministic Strategies \cite{cav-base-oop}}
\label{sec:oop-np}

We now consider a version of the optimal observability problem in which only positional and deterministic strategies are taken into account, making it decidable.
Recall that a positional and deterministic strategy for $\pomdp$ assigns one action to every state, \ie it is of the form $\sigma\colon S_\pomdp \to \Act_\pomdp$.
Formally, let $\OStratPD(\pomdp)$ denote the set of all positional and deterministic strategies for $\pomdp$. The minimal expected reward over such strategies $\MinExpRewPD{\pomdp}$ is determined by \Cref{def:min-exp-rew} with $\chi = \OStratPD(\pomdp)$.
The \emph{positional deterministic optimal observability problem} (\textbf{PDOOP}) is then defined as in \Cref{def:oo-problem}, but using $\MinExpRewPD{\pomdp}$ instead of $\MinExpRew{\pomdp}$. Konsta et al. \cite{cav-base-oop} conclude that the \OOP becomes \textsc{np}-complete when restricted to positional \emph{and} deterministic strategies.

\begin{example}[ctnd.]
Consider the \textbf{PDOOP}-instance $(\mdpline,2,4)$, where $\mdpline$ is the MDP in \Cref{POMDPprobs} for $p = \nicefrac{1}{2}$.
Then there is a solution by assigning the observation $o_1$ to $s_0$ and $s_1$ (and moving $r$(ight) for $o_1$), and $o_2$ to $s_3$ and $s_4$ (and moving $\ell$(eft) for $o_2$).
\end{example}

We remark that for positional and deterministic strategies, we can solve a stronger problem than optimal observability: how many observables (minimal budget) are required to turn an MDP into POMDP that yields the same minimal expected reward?

\begin{definition}[Minimal Positional Budget Problem (MPBP)]\label{def:minimalBudget}
  Given an MDP $\mdp$, determine an observation function 
  $\obs\colon S_\mdp \to \OO$ such that
  \vspace{-0.2cm}
  \begin{itemize}
  	\item $\MinExpRewPD{\mdp\setObs{\obs}} = \MinExpRewPD{\mdp}$ and
  	\item $\MinExpRewPD{\mdp\setObs{\obs}} < \MinExpRewPD{\mdp\setObs{\obs'}}$ for all observation functions $\obs'\colon S_\mdp \to \OOp$ with $|O'| < |O|$.
  \end{itemize}
  We call the number of observations $|O|$ the \emph{minimal positional budget} (\textbf{MPB}), and interchangeably refer to it as $B^{*}$.
\end{definition}
%


%
\subsection{Optimal Observability for Positional and Randomised Strategies \cite{cav-base-oop}}
\label{sec:oop-pos-randomised}
Going forward, we consider the \OOP for positional and \emph{possibly randomised} strategies, such that the admitted observable budget is lower than the minimal required as established in \Cref{def:minimalBudget}. An approach for randomised strategies was presented in \cite{cav-base-oop} and builds upon a typed extension of parameter synthesis techniques for Markov chains.
For a comprehensive overview of parameter synthesis techniques, we refer to~\cite{junges2020parameter,jansen2022-parameter-synthesis-survey}. 

\smallskip
\noindent\textbf{Typed Parametric Markov Chains \cite{cav-base-oop}.}
A typed parametric Markov chain (tpMC) admits expressions instead of constants as transition probabilities. The model admits variables (also called \emph{parameters}) of different types in expressions.
The types $\mathbb{R}$ and $\mathbb{B}$ represent real-valued and $\{0,1\}$-valued variables, respectively.
We denote by $\mathbb{R}_{=C}$ (resp. $\mathbb{B}_{=C}$) a type for real-valued (resp. $\{0,1\}$-valued) variables such that the values of all variables of this type sum up to some fixed constant $C$.\footnote{We allow using multiple types with different names of this form. For example, $\mathbb{R}^{1}_{= 1}$ and $\mathbb{R}^{2}_{= 1}$ are types for two different sets of variables whose values must sum up to one.}
Furthermore, we denote by $V(T)$ the subset of $V$ consisting of all variables of type $T$.
Moreover, $\Poly{V}$ is the set of multivariate polynomials with rational coefficients over variables taken from $V$.
\begin{definition}[Typed Parametric Markov Chains]\label{def:tpmc}
  A \emph{typed parametric Markov chain} is a tuple 
  $\pmc = (S, I, G, V, P, \rew)$, where $S$ is a finite set of \emph{states}, 
  $I \subseteq S$ is a set of \emph{initial states}, 
  $G \subseteq S$ is a set of \emph{goal states}, 
  $V$ is a finite set of typed variables, 
  $P\colon S \times S \rightarrow \Poly{V}$ is a \emph{parametric transition probability function}, and
  $\rew\colon S \rightarrow \mathbb{R}_{\geq 0}$ is a \emph{reward function}.
\end{definition}

\medskip
\medskip
\medskip
\noindent An \emph{instantiation} of a tpMC $\pmc$ is a function $\iota\colon V_\pmc \to \mathbb{R}$ such that
\vspace{-0.3cm}
\begin{itemize}
	\item for all $x \in V_\pmc(\mathbb{B}) \cup V_\pmc(\mathbb{B}_{= C})$, we have $\iota(x) \in \{0,1\}$;
	\item for all $V_\pmc(\mathbb{D}_{= C}) = \{ x_1,\ldots,x_n \} \neq \emptyset$ with $\mathbb{D} \in \{\mathbb{B},\mathbb{R}\}$, we have $\sum_{i=1}^{n} \iota(x_i) = C$. 
\end{itemize}
Given a polynomial $q \in \Poly{V_{\pmc}}$, we denote by $\instance{q}{\iota}$ the real value obtained from replacing in $q$ every variable $x \in V_{\pmc}$ by $\iota(x)$.
We lift this notation to transition probability functions by setting $\instance{P_\pmc}{\iota}(s,s') = \instance{P_\pmc(s,s')}{\iota}$ for all states $s, s' \in S_\pmc$.
An instantiation $\iota$ is \emph{well-defined} if it yields a well-defined transition probability function, i.e. if $\sum_{s' \in S_\pmc} \instance{P_\pmc}{\iota}(s,s') = 1$ for all $s \in S_\pmc$.
Every well-defined instantiation $\iota$ induces a Markov chain $\instance{\pmc}{\iota} = (S_\pmc, I_\pmc, G_\pmc, \instance{P}{\iota}, \rew_\pmc)$.

\begin{definition}[Feasibility Problem for tpMCs]\label{def:feasibility-tpmcs}
  Given a tpMC $\pmc$ and a threshold $\tau \in \mathbb{Q}_{\geq 0}$, 
  does there exist a well-defined instantiation $\iota$ such that 
  $\ExpRew{\instance{\pmc}{\iota}}{} \leq \tau$?
\end{definition}
Junges~\cite{junges2020parameter} studied decision problems for parametric Markov chains (pMCs) over real-typed variables.
In particular, he showed that the feasibility problem for pMCs over real-typed variables is \textsc{etr}-complete. Here, ETR refers to the \emph{Existential Theory of Reals}, \ie all true sentences of the form $\exists x_1 \ldots \exists x_n . P(x_1,...,x_n)$, where $P$ is a quantifier-free first-order formula over (in)equalities between polynomials with real coefficients and free variables $x_1, \ldots, x_n$. The complexity class \textsc{etr} consists of all problems that can be reduced to the \textsc{etr} in polynomial time.
Konsta et al. \cite{cav-base-oop} extend this result to show that the feasibility problem for tpMCs is \textsc{etr}-complete. Since \textsc{etr} lies between $\textsc{np}$ and $\textsc{pspace}$ (cf. \cite{DBLP:conf/stoc/Canny88}), it is also decidable in \textsc{pspace}.

\subsubsection{Positional Observability Problem}\label{POP}
The work in \cite{cav-base-oop} proved the optimal observability problem over positional strategies decidable. Formally, let $\OStrat_p(\pomdp)$ denote the set of all positional strategies for $\pomdp$. The minimal expected reward over such strategies $\MinExpRewP{\pomdp}$ is determined by \Cref{def:min-exp-rew} with $\chi = \OStratP(\pomdp)$.
\begin{definition}[Positional Observability Problem (POP)]\label{def:pos-observ-problem}
  Given an MDP $\mdp$, a budget $B \in \mathbb{N}_{\geq 1}$, and a threshold $\tau \in \mathbb{Q}_{\geq 0}$,
  is there a function $\obs\colon S_\mdp \to \OO$ with $|O| \leq B$ such that $\MinExpRewP{\mdp\setObs{\obs}} \leq \tau$?
\end{definition}
%
To solve a \textbf{POP}-instance $(\mdp,B,\tau)$, a tpMC $\pmc$ is constructed such that every well-defined instantiation corresponds to an induced Markov chain $\mdp\setObs{\obs}[\sigma]$ obtained by selecting an observation function $\obs\colon S_\mdp \to \{1,\ldots,B\} \uplus \{\obsGoal\}$ and a positional strategy $\sigma$.
Then, the \textbf{POP}-instance $(\mdp,B,\tau)$ has a solution iff the feasibility problem for $(\pmc, \tau)$ has a solution, as per \cite{cav-base-oop}.
Positional randomised POMDP strategies consider each action $\alpha$ with some \emph{unknown} probability depending on the given observation $o$, which is encoded by typed parameters $x_{o, \alpha}$.
Those parameters must form a probability distribution for every observation $o$, i.e. they will be of type $\mathbb{R}^{o}_{=\,1}$.
In the transition probability function, we then pick each action with the probability given by the parameter for the action and the current observation.
To encode the observation function $\obs$, we introduce a Boolean variable $y_{s,o}$ for every state $s$ and observation $o$ that evaluates to $1$ iff $\obs(s) = o$.
Formally, the tpMC $\pmc$ is constructed as follows:
\begin{definition}[Observation tpMC of an MDP]\label{def:observ-tpmc}
For an MDP $\mdp$ and a budget $B \in \mathbb{N}_{\geq 1}$, the corresponding \emph{observation tpMC} $\mathcal{D}_{\mdp} = (S_\mdp, I_\mdp, G_\mdp, V, P, \rew_\mdp)$ is given by
\begin{align*}
  & O ~=~ \{1, \ldots, B\} 
  \hspace{36mm}
  V ~=~ \biguplus_{s \in S_\mdp \setminus G_\mdp} V(\mathbb{B}^{s}_{=1}) ~~\uplus~~ \biguplus_{o \in O} V(\mathbb{R}^o_{=\,1})
  \\
  & V(\mathbb{B}^{s}_{=1}) ~=~ \{  y_{s,o} \mid o \in O \} 
  \qquad\qquad
  V(\mathbb{R}^o_{=\,1}) ~=~ \{  x_{o,\alpha} \mid
   \alpha \in \Act_\mdp \}
  \\[0.25em]
   & 
   \hspace{3cm}P(s,s') ~=~ 
  	\sum\limits_{\alpha \in \Act_\mdp}  \sum\limits_{o \in O} y_{s,o} \cdot x_{o,\alpha} \cdot P_\mdp(s,\alpha)(s'),
\end{align*}
where, to avoid case distinctions, we define $y_{s,o}$ as the constant $1$ for all $s \in G_\mdp$.
\end{definition}
The construction is sound in the sense that every Markov chain obtained from an MDP $\mdp$ by selecting an observation function and an observation-based positional strategy corresponds to a well-defined instantiation of the observation tpMC of $\mdp$. This yields a decision procedure for the positional observability problem: Given a \textbf{POP}-instance ($\mdp,B,\tau)$, construct the observation tpMC $\pmc$ of $\mdp$ for budget $B$.

Since tpMC feasibility is decidable in \textsc{etr}, it is also decidable
 whether there exists a well-defined instantiation $\iota$ such that $\ExpRew{\pmc[\iota]}{} \leq \tau$. This holds, by soundness, iff there exists an observation function $\obs\colon S \to \{1,\ldots B\} \uplus \{\obsGoal\}$ and a positional strategy $\sigma \in \StratP(\mdp\setObs{\obs})$ such that $\MinExpRewP{\mdp\setObs{\obs}} \leq \ExpRew{\mdp\setObs{\obs}}{\sigma} \leq \tau$. Hence, we use from \cite{cav-base-oop} that:
\begin{theorem}\label{th:pop-problem}
The positional observability problem \textbf{POP} is decidable in \textsc{etr}.
\end{theorem}
\begin{example}[ctnd.] \label{ex:bellman}
\Cref{figure:b} depicts the observation tpMC of the MDP $\mdpline$ in \Cref{POMDPprobs} for $p = 1$ and budget $B = 2$. 
The Boolean variable $y_{s,o}$ is true if we observe $o$ for state $s$.
Moreover, $x_{o,\alpha}$ represents the rate of choosing action $\alpha$ when $o$ is been observed. 
As is standard for Markov models~\cite{DBLP:books/wi/Puterman94}, including parametric ones~\cite{junges2020parameter}, the expected reward can be expressed as a set of recursive Bellman equations (parametric in our case). For the present example those equations yield the following \textsc{etr} constraints:  
\[
\begin{array}{rcl}
r_0 & = & 1 + (y_{s_0,o_1} \cdot x_{o_1, \ell} + y_{s_0,o_2} \cdot x_{o_2, \ell}) \cdot r_0 
              + (y_{s_0,o_1} \cdot x_{o_1, r} + y_{s_0,o_2} \cdot x_{o_2, r}) \cdot r_1 \\
r_1 & = & 1 + (y_{s_1,o_1} \cdot x_{o_1, \ell} + y_{s_1,o_2} \cdot x_{o_2, \ell}) \cdot r_0 
              + (y_{s_1,o_1} \cdot x_{o_1, r} + y_{s_1,o_2} \cdot x_{o_2, r}) \cdot r_2 \\
r_2 & = & 0, ~~r_3 = \ldots, ~~r_4 = \ldots \\
\tau & \geq & \frac{1}{4} \cdot (r_0 + r_1 + r_3 + r_4) 
\end{array}
\]
\noindent where $r_i$ is the expected reward for paths starting at $s_i$, i.e. $r_i = \sum_{\pi \in \Paths{\mdpline} \mid \pi[0]=s_i} \Prob{\mdpline}{\sigma}{\pi} \cdot \Rew{\mdpline}{\pi}$. Note that $\ExpRew{\mdpline}{\sigma} =  \frac{1}{4} \cdot (r_0 + r_1 + r_3 + r_4)$ for the strategy $\sigma $ defined by the parameters $x_{o,\alpha}$. 
\end{example}
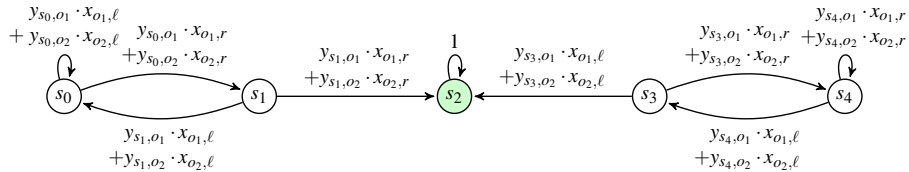
\begin{figure}[ht]
\vspace{-0.8cm}
\adjustbox{max width=\textwidth}{
\begin{tikzpicture}[->,>=stealth',shorten >=1pt,auto,node distance=3.0cm,
                    semithick]
  \tikzstyle{every state}=[fill=white,draw=black,text=black]

  \node[state,inner sep=2pt,minimum size=1pt]         (S0)                   {$s_0$};
  \node[state,inner sep=2pt,minimum size=1pt]         (S1) [right of=S0]     {$s_1$};
  \node[state,inner sep=2pt,minimum size=1pt][fill=green!20]          (S2) [right of=S1]     {$s_2$};
  \node[state,inner sep=2pt,minimum size=1pt]         (S3) [right of=S2]     {$s_3$};
  \node[state,inner sep=2pt,minimum size=1pt]         (S4) [right of=S3]     {$s_4$};

  \path (S0) edge   [bend left=20]           node [above][align=right]{\hspace{0.4cm}${\color{white}+}y_{s_0,o_1} \cdot x_{o_1, r}$\\$+y_{s_0,o_2} \cdot x_{o_2, r}$} (S1)
             edge   [loop above]             node [above][align=center] {${\color{white}+}\ y_{s_0,o_1} \cdot x_{o_1, \ell}$\\$+\ y_{s_0,o_2} \cdot x_{o_2, \ell}$ } (S0)
        (S1) edge                            node [above][align=center] {${\color{white}+}y_{s_1,o_1} \cdot x_{o_1, r}$\\$+y_{s_1,o_2} \cdot x_{o_2, r}$ } (S2)
             edge   [bend left=20]           node [below][align=center] {${\color{white}+}y_{s_1,o_1} \cdot x_{o_1, \ell}$\\$  +y_{s_1,o_2} \cdot x_{o_2, \ell}$ } (S0)
        (S2) edge   [loop above]             node [above]{$1$} (S2)
        (S3) edge   [bend left=20]           node [above][align=center] {\hspace{-0.3cm}${\color{white}+}y_{s_3,o_1} \cdot x_{o_1, r}$\\ \hspace{-0.3cm}$ +y_{s_3,o_2} \cdot x_{o_2, r}$ } (S4)
             edge                            node [above][align=center] {${\color{white}+}y_{s_3,o_1} \cdot x_{o_1, \ell}$\\$ +y_{s_3,o_2} \cdot x_{o_2, \ell}$ } (S2)
        (S4) edge   [bend left=20]           node [below][align=center] {${\color{white}+}y_{s_4,o_1} \cdot x_{o_1, \ell}$\\$ +y_{s_4,o_2} \cdot x_{o_2, \ell}$ } (S3)
             edge   [loop above]             node [above right=0cm and -0.8cm][align=center] {${\color{white}+}y_{s_4,o_1} \cdot x_{o_1, r}$\\$ +y_{s_4,o_2} \cdot x_{o_2, r}$ } (S4);
\end{tikzpicture}
}
  \caption{Observation tpMC for the MDP $\mdpline$ in~\Cref{POMDPprobs} with $p=1$ and budget 2.}
  \label{figure:b}
  \vspace{-0.5cm}
\end{figure}

\subsubsection{Sensor Selection Problem}\label{SSP}
We finally consider a variant of the positional observability problem in which observations can only be made through a fixed set of location sensors that can be turned on or off for every state.
The \emph{sensor selection problem} aims at turning an MDP into a location POMDP with a limited number of observations such that the expected reward stays below a threshold.
In this scenario, a POMDP can either observe its position (i.e. the current state) or nothing at all (represented by $\bot$).
Formally, we consider \emph{location POMDPs} $\pomdp$ with observations $O_\pomdp = D \uplus \{ \bot \}$, where $D  \subseteq \{ @s \mid s \in (S_\pomdp \setminus G_\pomdp) \}$ are the observable locations and the observation function is

\begin{align*}
  \obs_\pomdp(s) ~=~
  \begin{cases}
  	@s,   ~\text{if}~ @s \in D \\
  	\obsGoal, & ~\text{if}~ s \in G_\pomdp \\
  	\bot, & ~\text{if}~ @s \notin D ~\text{and}~ s \notin G_\pomdp .
  \end{cases}	
\end{align*}
\vspace{-0.2cm}
\begin{example}[ctnd.]
  Consider the MDP $\mdpline$ with $p=1$ and location sensors assigned as in \Cref{POMDPprobs}. With a budget of $2$ we can only select 2 of the 4 location sensors. For example, we can turn on the sensors on one side, say $@s_0$, $@s_1$.
    The observation function is then given by $\obs(s_0)=@s_1$, $\obs(s_1)=@s_2$, and $\obs(s_3)=\obs(s_4) = \bot$.
    This is an optimal sensor selection as it reveals whether one is located left or right of the goal. 
\end{example}
\begin{definition}[Sensor Selection Problem (SSP)]\label{def:sensor-select-problem} Given an MDP $\mdp$, a budget $B \in \mathbb{N}_{\geq 1}$, and $\tau \in \mathbb{Q}_{\geq 0}$,
  is there an observation function $\obs\colon S_\mdp \to \OO$ with $|O| \leq B$ such that $\pomdp = (\mdp,O,\obs)$ is a location POMDP and  $\MinExpRewP{\pomdp} \leq \tau$? 
\end{definition}
To solve the \textbf{SSP}, one constructs a tpMC similar to \Cref{def:observ-tpmc}.
The main difference is that we use a Boolean variable $y_i$ to model whether the location sensor $@s_i$ is on ($1$) or off ($0$).
Moreover, we explicitly require that at most $B$ sensors are turned on.
\begin{definition}[Location tpMC of an MDP]\label{def:location-tpmc}
For an MDP $\mdp$ and a budget $B \in \mathbb{N}_{\geq 1}$, the corresponding \emph{location tpMC} $\mathcal{D}_{\mdp} = (S_\mdp, I_\mdp, G_\mdp, V, P, \rew_\mdp)$ is given by
\begin{align*}
  V = V(\mathbb{B}_{= B}) \uplus \biguplus_{o \in O} V(\mathbb{R}^o_{=\,1})
  \quad  
  V(\mathbb{B}_{= B}) = \{  y_{s} \mid s \in S_\mdp \setminus G_\mdp \} 
  \quad 
  V(\mathbb{R}^o_{=\,1}) = \{  x_{s,\alpha} \mid \alpha \in \Act_\mdp \}
  \\
   P(s,s') ~=~ \sum\limits_{\alpha \in \Act}  
   y_s \cdot x_{s,\alpha}\cdot P(s,\alpha)(s') 
   +
   (1-y_s) \cdot x_{\bot,\alpha}\cdot P(s,\alpha)(s'),
   \hspace{3em}
\end{align*}
where, to avoid case distinctions, we define $y_s$ as the constant $1$ for all $s \in G_\mdp$.
\end{definition}
Analogous to observation tpMCs, the sensor selection problem admits a decision procedure in \textsc{pspace} due to the soundness argument for location tpMCs~\cite{cav-base-oop}.
\begin{theorem}\label{def:col-oops-problem}
The sensor selection problem \textbf{SSP} is decidable in \textsc{etr}, and thus in \textsc{pspace}.
\end{theorem}
\begin{example}
\Cref{figure:c} shows the location tpMC of the location POMDP in \Cref{POMDPprobs} for $p=1$ and budget $2$.
The Boolean variable $y_{s}$ indicates if the sensor $@ s$ is be turned on, while the variables $x_{s,\alpha}$ indicates the rate of choosing action $\alpha$ if sensor $@ s$ is turned on; otherwise, if sensor $@s$ is turned off, $x_{\bot,\alpha}$ is used, which is the rate of choosing action $\alpha$ for unknown locations.  
\end{example}
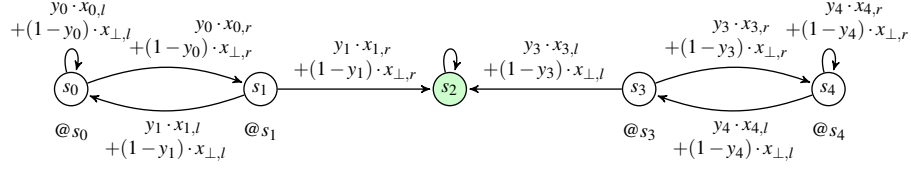
\begin{figure}[ht]
\vspace{-0.15cm}
\adjustbox{max width=\textwidth}{
	\begin{tikzpicture}[->,>=stealth',shorten >=1pt,auto,node distance=3cm,
                    semithick]
  \tikzstyle{every state}=[fill=white,draw=black,text=black]

  \node[state,inner sep=2pt,minimum size=1pt]         (S0)                   {$s_0$};
  \node[state,inner sep=2pt,minimum size=1pt]         (S1) [right of=S0]     {$s_1$};
  \node[state,inner sep=2pt,minimum size=1pt] [fill=green!20]         (S2) [right of=S1]     {$s_2$};
  \node[state,inner sep=2pt,minimum size=1pt]         (S3) [right of=S2]     {$s_3$};
  \node[state,inner sep=2pt,minimum size=1pt]         (S4) [right of=S3]     {$s_4$};

  \path (S0) edge   [bend left=20]           node [above][align=right]{\hspace{1.6cm}\footnotesize${\color{white}+}y_{0} \cdot x_{0, r}$\\ \footnotesize$ +  (1 - y_0) \cdot x_{\bot,r}$} (S1)
             edge   [loop above]             node [above][align=center] {\footnotesize${\color{white}+}y_{0} \cdot x_{0, l} $\\ \footnotesize$ +  (1 - y_0) \cdot x_{\bot,l}$ } (S0)
        (S1) edge                            node[above][align=center] {\footnotesize ${\color{white}+}y_{1} \cdot x_{1, r} $\\ \footnotesize$+  (1 - y_1) \cdot x_{\bot,r}$} (S2)
             edge   [bend left=20]           node [below][align=center]{\footnotesize${\color{white}+}y_{1} \cdot x_{1, l} $\\ \footnotesize$ +  (1 - y_1) \cdot x_{\bot,l}$} (S0)
        (S2) edge   [loop above]             node [above]{} (S2)
        (S3) edge   [bend left=20]           node [above][align=center]  {\footnotesize${\color{white}+}y_{3} \cdot x_{3, r}  $\\ \footnotesize$+  (1 - y_3) \cdot x_{\bot,r}{\color{white}+}$} (S4)
             edge                            node [above][align=center]{\footnotesize ${\color{white}+}y_{3} \cdot x_{3, l}  $\\ \footnotesize$+  (1 - y_3) \cdot x_{\bot,l}$ } (S2)
        (S4) edge   [bend left=20]           node [below][align=center] {\footnotesize${\color{white}+}y_{4} \cdot x_{4, l}  $\\ \footnotesize$+  (1 - y_4) \cdot x_{\bot,l}$ } (S3)
             edge   [loop above]             node [above right=0cm and -0.8cm][align=center]{\footnotesize${\color{white}+}y_{4} \cdot x_{4, r}  $\\ \footnotesize$+  (1 - y_4) \cdot x_{\bot,r}$} (S4);
\node[below=0.15cm of S0] (s1) {$@ s_0$};
  \node[below=0.15cm of S1] (s2) {$@ s_1$};
  \node[below=0.15cm of S3] (s3) {$@ s_3$};
  \node[below=0.15cm of S4] (s4) {$@ s_4$};

\end{tikzpicture}
}
\caption{Location tpMC for the location POMDP in~\Cref{POMDPprobs} with $p=1$ and budget $2$.}
\label{figure:c}
\vspace{-0.5cm}
\end{figure}

\vspace{-0.2cm}
\section{Enhancements of SMT-based Parameter Synthesis}
\label{sec:enhancements-smt}
Previous research \cite{cav-base-oop} introduced $2$ methods for solving \OOP instances: (a) brute-force enumeration of observation functions and strategies combined with PRISM model checking, and (b) a symbolic approach that encodes problems into \textsc{etr}-systems to be solved using SMT. A detailed formalisation of the constraints used in the SMT parameter synthesis is provided in \cite{msc-thesis}.
While standard parameter synthesis tools are restricted to graph-preserving models and real-valued variables, the tpMC structure overcomes these topology limits and achieves significantly better performance through SMT-backed queries rather than PRISM enumeration, especially for reward thresholds that approach the optimum \cite{cav-base-oop}.

Motivated by this superior performance, we attempt to exploit the full potential of SMT parameter synthesis through a series of refinements (\Cref{subsec:smt-constraint-experiments,subsec:bool-encoding,subsec:relaxations}) for the \textsc{etr}-encodings that \cite{cav-base-oop} proposes. 
We reproduced the baseline environment and results from \cite{cav-base-oop} using \zthree version $\texttt{4.13.0}$, considering that later releases reduce the performance of the SMT problems in this work.
\Cref{subsec:pomdp-oracle} presents the combined results for our best-performing enhancements, which were evaluated using the same benchmarks as \cite{cav-base-oop} and beyond on the following specifications: Apple M2 processor chip, 8 GB RAM. To ensure reproducible benchmarking for our experiments, we have containerised these workflows and exposed solver configurations through runtime parameters. We confirmed the reliability of our further results by assessing a negligible coefficient of variation in CPU runtime across multiple benchmark trials.

\subsection{SMT Constraint Experiments}\label{subsec:smt-constraint-experiments}
A known limitation of SMT solvers manifests when small semantic-preserving changes, such as reordering assertions, symbol renaming, and adding or removing redundant variables, lead to significant variations in results or performance. In the literature, this issue is called the \emph{(in)stability problem}, and is caused by the complex heuristics that drive the solvers. It has been reported extensively \cite{komodo-smt-instability,dafny-smt-instability,ironclad-smt-instability}, and there have been attempts to measure and mitigate it \cite{mitigate-smt-instability,measure-smt-instability}.
%
%
We study the problem with independent experiments in an attempt to improve our solver’s performance. 

We explore alternatives to the arbitrary assertion order of SMT constraints that was chosen in \cite{cav-base-oop}, as well as equivalent formulations of the Bellman equations.
Specifically, to manage the computational cost of this exploration, we grouped SMT constraints into four distinct categories based on their semantic role in the \POP- and \SSP-encodings. We then evaluated their permutations, with the internal order within each category fixed. 
The empirical results demonstrate that indeed the solver performance is highly sensitive to assertion order, with variations in runtime above $25\%$. Despite this fact, the original order proved to dominate the average solving time and solvability of all category permutations, and was hence selected for the rest of the work.
Similarly, a common finding is that the refutation time, \ie, the processing time to determine that an \OOP query is unsatisfiable, was not susceptible to such effects.
For the curious reader, a detailed presentation of the experiments and results is offered in \cite{msc-thesis}.

\subsection{Native Boolean Encoding}\label{subsec:bool-encoding}
Since the original \cite{cav-base-oop} tpMC implementation in SMT used exclusively $\texttt{Real}$-typed parameters restricted to the binary domain, we substituted them with a native Boolean encoding that leverages SAT propagation.
In \Cref{def:tpmc}, the implementation from \cite{cav-base-oop} corresponds to an empty set of Boolean variables $V_\pmc(\mathbb{B}) = V_\pmc(\mathbb{B}_{=C}) = \emptyset$ as it implemented extensions of \Cref{def:observ-tpmc} with $V(\mathbb{R}^s_{=1}) = \{y_{s,o} \mid o \in O\}$ and of \Cref{def:location-tpmc} with $V(\mathbb{R}_{=B})  = \{y_s \mid s \in S_M \setminus G_M\}$, respectively.
We aim to minimise the non-linear arithmetic overhead in the tpMC constraint system and, in this way, enforce a tighter correspondence to \Cref{def:observ-tpmc} and \Cref{def:location-tpmc} in which variables denoting the observation assignments and deterministic strategies behave as native Booleans:
\[
\centering
V(\mathbb{B}^s_{=1}) = \{y_{s,o} \mid o \in O\},
~V(\mathbb{B}_{=B}) = \{y_s \mid s \in S_M \setminus G_M\},
~\text{and}~
V(\mathbb{B}^o_{=1}) = \{x_{o,a} \mid a \in Act\}
\]
This required reformulating typing constraints for observation assignments $y_{s,o}$ and deterministic strategies $x_{o,a}$ as logical cardinality clauses:
\[
\centering
\forall_{s \in S \setminus G}\colon\bigvee\limits_{o \in O} y_{s,o}
~~\text{\emph{and}}~~
\forall_{s \in S \setminus G}\colon \forall_{o \in O}\colon y_{s,o} \rightarrow \bigwedge\limits_{o' \in O \setminus \{o\}} \neg y_{s, o'}
\]
Accordingly, we have restructured the Bellman equations to aggregate future rewards through logical implications, and implemented the budget constraint as a sum of conditionals. For a detailed formalisation of these constraints, we refer to \appref{appendix:bool-enc}. 

For deterministic strategies, we observed substantial improvements. While isolated instances exhibited similar gains for randomised strategies, the performance could not generalise to the entire benchmark. To further improve the runtime performance for randomised strategies, we use Pseudo-boolean cardinality constraints for sensor budget and typing of observation function parameters. This approach relies extensively on a dedicated solver within \zthree that, even when mixed with the non-linear real arithmetic theory, significantly speeds up the solver. This encoding has performed the bulk of the speedup for the benchmark results detailed in \Cref{subsec:pomdp-oracle}.
\vspace{-0.3cm}

\subsection{Relaxations}\label{subsec:relaxations}

Following, we discuss a couple of problem relaxations in an attempt to further improve the solver performance.

\subsubsection{Budget Repairing in a Multiple-shot Approach:} 
We explored an incremental solving strategy inspired by stack-based (push/pop) SMT interactions, proposing a two-stage ``relax-and-repair'' procedure:

\begin{enumerate}
    \vspace{-0.3cm}
    \item \textbf{relaxation} - solver computes a valid solution for the unconstrained problem which effectively ignores the budget constraint of SSP.
    \item \textbf{repair} - the budget constraint is pushed onto the solver stack, and solver attempts to refine the current state into a compliant solution.
    \vspace{-0.2cm}
\end{enumerate}

We hypothesised that the solver could leverage cached internal state, such as learned clauses and conflict analysis from a relaxed phase, to locally ``repair'' the solution rather than initiate a global search. When applied to the original real encoding, the technique delivered significant performance gains, reducing solving times by $\sim 15$ times. The optimisation failed to transfer to the native Boolean reformulation.
%

\subsubsection{Invariant Upper Bound on Rewards (Bellman inequalities):}
We introduced a logical relaxation of the reward formulation in our SMT encodings, which replaces standard Bellman equations with inequalities $v_s \geq Bellman(v_s)$. This formulation leverages the fact that our parameter synthesis objective is bounded on the given threshold rather than optimality. Rather than solving for the exact expected reward, we seek, in theory, a conservative over-approximation of the true expected reward towards the goal from each state. It leads to the invariant property: if the upper bound satisfies the reward threshold $\tau$, the actual minimal cost $\MinExpRew{\pomdp}$ necessarily satisfies it as well:
\vspace{-0.2cm}
\[
\MinExpRew{\pomdp} \leq \frac{1}{\iota}\sum_{x\in S \setminus G}v_x \leq \tau
\]
This relaxation admits looser guarantees on feasible rewards, \ie not necessarily equal to the truthful expected reward, whenever the acceptance criterion is set above the optimal expected reward $\tau > \MinExpRew{M}$. Note that the higher the distinguishability power (budget) for an observation function $\obs$, the lower the minimal expected reward of a POMDP $\mdp\setObs{\obs}$ (see \cite{msc-thesis} for a proof of the argument). We relaxed the budget constraint to an ``at most $k$" formulation suitable for arithmetic optimisation, that deliberately ignores the reward monotonicity w.r.t. information.

This relaxation proved highly effective for the original $\texttt{Real}$-typed encoding, reducing solving times by up to $8$ times for $6/18$ instances in the corresponding benchmark. The additional budget relaxation, had a positive effect for $4$ more instances. For the $\texttt{Boolean}$-typed encoding, the approach performed best when maintaining the strict budget equality for a tighter search space and helped lower the solving time of both deterministic and randomised strategies.


\subsection{Experimental Results}\label{subsec:pomdp-oracle}

\newcommand{\timeout}[1]{\cellcolor{black!30} t.o.}
\newcommand{\na}[1]{N/A}
\newcommand{\nato}[1]{\cellcolor{black!30} N/A}
\newcommand{\unknown}[1]{\cellcolor{white} unk}
\newcommand{\fail}[1]{\cellcolor{black!30}}
\newcommand{\heaps}[1]{\cellcolor{black!30} o.o.m.}
\newcommand{\improve}[1]{\cellcolor{ForestGreen!50} #1}
\renewcommand{\arraystretch}{1.3}

\Cref{tab:smt-ran-results,tab:smt-det-results} compare the performance of our combined refinements, namely the native boolean encoding and invariant relaxation, against the largest solvable instances presented in the original approach \cite{cav-base-oop}. The left- and right-hand side of the tables shows the results for \textbf{P(DO)OP} and \SSP, respectively. We benchmark the same problem instances presented in \cite{cav-base-oop}, which are standard models in the POMDP literature (grid(world)~\cite{LITTMAN1995362} and maze~\cite{MCCALLUM1993190}, and of the $\mdpline$ example in \Cref{figure:b}). The included variations are queried with various state space sizes, budgets and thresholds, namely $\tau\colon \leq 2\cdot\MinExpRew{M}$, $\leq \MinExpRew{M}$ and $< \MinExpRew{M}$, where $\mdp$ is the instance's underlying MDP.
We denote by $\text{L}(k)$ a variant of the MDP $\mdpline$ scaled up to $k$ states. $\text{G}(k)$ is a $k \times k$ grid model where the goal state is in the bottom-right corner. Finally, $\text{M}(k)$ refers to the maze model, where $k$ is the (odd) number of states. An example of the maze can be seen in \Cref{app:mazeExample}.

The results presented are strictly for the feasibility formulation of such problems (cf. \Cref{def:feasibility-tpmcs}) and reflect budgets equal to the \textbf{MPB}. An attentive reader will notice that optimal observability with positional and deterministic strategies admits no solutions when the queried budget $B < B^{*}$. In all but one case, the reward computed by \zthree (if any) is the same as the optimal $\MinExpRew{M}$. The column Original(\zthree) indicates the runtime result in \cite{cav-base-oop}, and Current(\zthree) indicates the performance of our selected collection of refinements (original assertion order, with Boolean formulation and invariant relaxation). We write t.o. if a variant exceeds the timeout of $3$ minutes, and highlight in green the best performing version, if any of the two succeed. The source code provides an extensive repository of benchmark outcomes for independent experiments and various combinations of refinements, which did not scale to the same extent and performance.

\Cref{tab:smt-ran-results,tab:smt-det-results} show that the approaches investigated in this section improved the runtime by many orders of magnitude. Only $\text{G}(20)$ with threshold $\frac{7600}{399}$ (cf. \Cref{tab:smt-ran-results}) solved slower than the original results, a behavior we attribute to the typical instability of \zthree. Furthermore, we can observe larger improvements in runtime performance for deterministic strategies than for randomized ones. Boolean encoding accounts for much of this difference, since the binarisation of strategy constraint variables in deterministic strategies implies one can faithfully encode those variables using \texttt{Boolean}-typed parameters and reduce the number of constraints asserted to the SMT solver. The performance difference among the considered strategies is also evident in their sensitivity to thresholds: larger thresholds lead to substantially longer solver runtime (or timeouts) for randomised strategies, but not for deterministic ones. In the original version, this effect was noticeable for both strategies. 
\Cref{tab:smt-results-scaled} compares the largest solvable instance dimensions for each considered problem variant and world topology, measured in the number of states $|S_\mdp|$ of the $(\mdp, B, \tau)$ instances.
Current enhancements for the SMT parameter synthesis achieve a higher dimensionality than the original implementation with scaling factors ranging between $1$ and $\sim 77$. Once again, we observe that deterministic strategies achieve superior results in terms of instance size.
%

\begin{table}[ht!]
\vspace{-0.5cm}
\centering
\scriptsize
\begin{minipage}[t]{0.49\textwidth}
\begin{tabular}{|l|l|l|C{1.65cm}|C{1.3cm}|}
 
 \multicolumn{5}{c}{\POP - Randomised Strategies} \\
 \hline
 \multicolumn{3}{|c|}{Problem Instance} & Original(\zthree)\cite{cav-base-oop} & Current(\zthree) \\
 \hline
 Model & Thresh. & Budget & Time(s) & Time(s) \\
 \hline \hline

   \multirow{3}{*}{L$(249)$} &  $\leq \frac{250}{2}$  & 2  & \timeout{} & \timeout{} \\ \cline{2-5}
  & $ \leq \frac{125}{2}$ & 2 & $19.051$  & \improve{$0.2715$} \\ \cline{2-5}
  & $< \frac{125}{2} $ & 2 & $15.375$ & \improve{$0.0254$} \\

\hline\hline

  \multirow{3}{*}{G$(20)$} &  $\leq \frac{15200}{399}$ & 2  & \timeout{}  & \timeout{} \\ \cline{2-5}
  &  $\leq \frac{7600}{399}$ & 2  & \improve{$19.164$}  & $30.896$ \\ \cline{2-5}
  &  $< \frac{7600}{399}$    & 2  & $15.759$  & \improve{$0.0502$} \\ 
  
  \hline\hline

 \multirow{3}{*}{M$(7)$} &  $\leq \frac{168}{15}$  & 4  & \timeout{}  & \timeout{} \\ \cline{2-5}
 &  $\leq \frac{84}{15}$  & 4  & $15.598$  & \improve{$0.1849$} \\ \cline{2-5}
  &  $< \frac{84}{15}$     & 4  & $31.986$ & \improve{$0.0082$} \\ 
 \hline

\end{tabular}
\end{minipage}
\scriptsize
\hfill
\begin{minipage}[t]{0.49\textwidth}
\begin{tabular}{|l|l|l|C{1.65cm}|C{1.3cm}|}
 
 \multicolumn{5}{c}{\SSP - Randomised Strategies} \\
 \hline
 \multicolumn{3}{|c|}{Problem Instance} & Original(\zthree)\cite{cav-base-oop} & Current(\zthree) \\
 \hline
 Model & Thresh. & Budget & Time(s) & Time(s) \\
 \hline \hline

   \multirow{3}{*}{L$(61)$} &  $\leq 31$  & 30  & \timeout{} & \improve{$0.3819$} \\ \cline{2-5}
  & $ \leq \frac{31}{2}$ & 30 & $17.894$ & \improve{$0.0215$} \\ \cline{2-5}
  & $< \frac{31}{2} $ & 30 & $30.198$ & \improve{$0.0107$} \\

\hline\hline

   \multirow{3}{*}{G$(6)$} &  $\leq \frac{360}{35}$ & 5  & \timeout{} & \timeout{}  \\ \cline{2-5}
  &  $\leq \frac{180}{35}$ & 5  & $16.671$ & \improve{$0.1831$} \\ \cline{2-5}
  &  $< \frac{180}{35}$    & 5  & $30.204$  & \improve{$0.0102$} \\

  \hline\hline

 \multirow{3}{*}{M$(15)$} &  $\leq \frac{868}{35}$  & 21  & \timeout{} & \timeout{} \\ \cline{2-5}
 &  $\leq \frac{434}{35}$  & 21  & $19.067$  & \improve{$3.4164$} \\ \cline{2-5}
  &  $< \frac{434}{35}$     & 21  & $30.463$ & \improve{$0.0099$} \\ 
 \hline

\end{tabular}
\end{minipage}
\vspace{0.1cm}
\caption{Comparison of experimental results for randomised strategies after SMT enhancements on Parameter Synthesis.}
\label{tab:smt-ran-results}
\vspace{-1cm}
\end{table}

\begin{table}[ht!]
\vspace{-0.3cm}
\centering
\scriptsize
\begin{minipage}[t]{0.49\textwidth}
\begin{tabular}{|l|l|l|C{1.65cm}|C{1.3cm}|} 
 \multicolumn{5}{c}{\textbf{PDOOP} - Deterministic Strategies} \\
 \hline
 \multicolumn{3}{|c|}{Problem Instance} & Original(\zthree)\cite{cav-base-oop} & Current(\zthree) \\
 \hline
 Model & Thresh. & Budget & Time(s) & Time(s) \\
 \hline \hline
  

  \multirow{3}{*}{L$(377)$} &  $\leq 189$  & 2  & $55.735$  & \improve{$0.0368$} \\ \cline{2-5}
  &  $\leq \frac{189}{2}$  & 2  & $19.148$  & \improve{$0.0367$}  \\ \cline{2-5}
  &   $< \frac{189}{2}$    & 2  & $353.311$ & \improve{$0.0344$} \\   
  
  \hline\hline

  \multirow{3}{*}{G$(24)$} &  $\leq \frac{26496}{575}$ & 2  & \timeout{}  & \improve{$0.0824$} \\ \cline{2-5}
  &  $\leq \frac{13248}{575}$ & 2  & $19.751$  & \improve{$0.0838$} \\ \cline{2-5}
  &  $< \frac{13248}{575}$    & 2  & $30.843$  & \improve{$0.0764$} \\  \hline\hline
 
  \multirow{3}{*}{M$(39)$} &  $\leq \frac{6232}{95}$  & 4  & \timeout{} & \improve{$0.0265$} \\ \cline{2-5}
 &  $\leq \frac{3116}{95}$  & 4  & $20.424$ & \improve{$0.0262$} \\ \cline{2-5}
  &  $< \frac{3116}{95}$     & 4  & $30.149$ & \improve{$0.0227$} \\ 
 \hline
\end{tabular}
\end{minipage}
\hfill
\begin{minipage}[t]{0.49\textwidth}
\begin{tabular}{|l|l|l|C{1.65cm}|C{1.3cm}|} 
 \multicolumn{5}{c}{\SSP - Deterministic Strategies} \\
 \hline
 \multicolumn{3}{|c|}{Problem Instance} & Original(\zthree)\cite{cav-base-oop} & Current(\zthree) \\
 \hline
 Model & Thresh. & Budget & Time(s) & Time(s) \\
 \hline \hline
  
  
  \multirow{3}{*}{L$(193)$} &  $\leq 97$  & 96  & \timeout{}   & \improve{$0.0206$} \\ \cline{2-5}
  &  $\leq \frac{97}{2}$  & 96  & $20.530$  & \improve{$0.0208$} \\ \cline{2-5}
  &   $< \frac{97}{2}$    & 96  & $30.412$ & \improve{$0.0184$} \\  

  \hline\hline

   \multirow{3}{*}{G$(15)$} &  $\leq \frac{3150}{112}$ & 14  &  \timeout{}  & \improve{$0.0415$} \\ \cline{2-5}
  &  $\leq \frac{3150}{224}$ & 14  & $20.204$   & \improve{$0.0413$} \\ \cline{2-5}
  &  $< \frac{3150}{224}$    & 14  & $30.804$  & \improve{$0.0375$}\\ 

  \hline\hline

 \multirow{3}{*}{M$(49)$} &  $\leq \frac{9912}{120}$  & 72  & \timeout{} & \improve{$0.0226$} \\ \cline{2-5}
 &  $\leq \frac{4956}{120}$  & 72  & $20.35$  & \improve{$0.0213$} \\ \cline{2-5}
  &  $< \frac{4956}{120}$ & 72  & $30.333$ & \improve{$0.0197$}\\ 

 \hline
\end{tabular}
\end{minipage}
\vspace{0.1cm}
\caption{Comparison of experimental results for deterministic strategies after SMT enhancements on Parameter Synthesis.}
\label{tab:smt-det-results}
\vspace{-0.2cm}
\end{table}

\begin{table}[ht!]
\vspace{-0.2cm}
\centering
\scriptsize
\begin{minipage}[t]{0.24\textwidth}
\begin{tabular}{|c|c|c|} 
 \multicolumn{3}{c}{\POP\ - Rand. Strategies} \\
 \hline
 \multirow{2}{*}{World} & \cite{cav-base-oop} & Current \\ \cline{2-3}
  & $\lvert S_M \rvert$ & $\lvert S_M \rvert$ \\ \hline \hline

  \Line & $249$ & \improve{$1~001$} \\ \hline
  \Grid & $400$ & $400$ \\ \hline
  \Maze & $16$ & \improve{$266$} \\ \hline
\end{tabular}
\end{minipage}
\hfill
\begin{minipage}[t]{0.24\textwidth}
\hspace{0.12cm}
\begin{tabular}{|l|c|c|} 
 \multicolumn{3}{c}{\SSP\ - Rand. Strategies} \\
 \hline
 \multirow{2}{*}{World} & \cite{cav-base-oop} & Current \\ \cline{2-3}
  & $\lvert S_M \rvert$ & $\lvert S_M \rvert$ \\ \hline \hline
  \Line & $61$ & \improve{$701$} \\ \hline
  \Grid & $36$ & \improve{$256$} \\ \hline
  \Maze & $36$ & \improve{$196$} \\ \hline
\end{tabular}
\end{minipage}
\hfill
\begin{minipage}[t]{0.24\textwidth}
\hfill
\begin{tabular}{|l|c|c|} 
 \multicolumn{3}{c}{\textbf{PDOOP} - Det. Strategies} \\
 \hline
 \multirow{2}{*}{World} & \cite{cav-base-oop} & Current \\ \cline{2-3}
  & $\lvert S_M \rvert$ & $\lvert S_M \rvert$ \\ \hline \hline

  \Line & $377$ & \improve{$10~001$} \\ \hline
  \Grid & $96$ & \improve{$1~296$} \\ \hline
  \Maze & $576$ & \improve{$846$} \\ \hline
\end{tabular}
\end{minipage}
\hfill
\begin{minipage}[t]{0.24\textwidth}
\hfill
\begin{tabular}{|l|c|c|} 
 \multicolumn{3}{c}{\SSP\ - Det. Strategies} \\
 \hline
 \multirow{2}{*}{World} & \cite{cav-base-oop} & Current \\ \cline{2-3}
  & $\lvert S_M \rvert$ & $\lvert S_M \rvert$ \\ \hline \hline

  \Line & $193$ & \improve{$15~001$} \\ \hline
  \Grid & $225$ & \improve{$10~000$} \\ \hline
  \Maze & $36$ & \improve{$2~621$} \\ \hline
\end{tabular}
\end{minipage}
\vspace{0.1cm}
\caption{Largest solvable instances after SMT enhancements (Current) on Parameter Synthesis (\cite{cav-base-oop}), measured in the number of states $\lvert S_\mdp \rvert $ of the $(\mdp, B, \tau)$ instances.}
\label{tab:smt-results-scaled}
\vspace{-0.6cm}
\end{table}

\section{Integrations for POMDP Analysis}\label{sec:integration-pomdp-oracles}

This section explores two alternative approaches towards efficiently evaluating POMDPs using SMT (\Cref{subsec:smt-oracle}) and Probabilistic Model Checking tools (\Cref{subsec:integration-storm}) such that they aid in solving \OOP variants. It lays the foundation for \Cref{sec:informed-search-algorithms} and preliminary research into Reinforcement Learning methods (cf. \cite{msc-thesis}), which study how to synthesise sensible observation functions. This constitutes a shift from the approach of \cite{cav-base-oop}: we no longer investigate how to solve the entire \tpmc encoding of \POP- and \SSP-instances, but rather focus on searching the space of observation functions and evaluating them using so-called POMDP ``oracles''. We denote by \emph{oracle} any procedure that evaluates a candidate solution to \POP/\SSP, namely a POMDP $\pomdp=\mdp\setObs{obs}$ induced by an instantiated observation function $\obs$ and MDP $\mdp$.

\subsection{SMT Oracle}\label{subsec:smt-oracle}
For a straightforward attempt to develop an oracle, we noticed potential in POMDP evaluation using a \tpmc encoding given the performance improvements of SMT-based parameter synthesis in \Cref{sec:enhancements-smt}.
To evaluate $\MinExpRewP{\pomdp}$ of a POMDP $\pomdp = (\mdp, O, \obs) = \mdp\setObs{\obs}$, we construct a \tpmc $\pmc_\pomdp = \pmc_{\mdp\setObs{\obs}}$ such that every well-defined instantiation corresponds to an induced Markov chain $\mdp\setObs{\obs}[\sigma]$ obtained by selecting a positional strategy $\sigma$. Then, the \POP- or \SSP-instance $(\mdp, B, \tau)$ has the certificate $\obs$ (and a solution) iff the feasibility problem for $(\pmc_{\mdp\setObs{\obs}}, \tau)$ has a solution.

We construct $\pmc_{\mdp\setObs{\obs}}$ from $\pmc_\mdp = (S_\mdp, I_\mdp, G_\mdp, V, P, \rew_\mdp)$ by fixing a value to the $y_{s,o}$ parameters according to the observation assignment $\obs$. More specifically, we substitute parameters in $V$ by applying the substitutions $\{y_{s,o}\mapsto \left ( \obs(s)=o \right )\}, \forall s\in S_M\colon \forall o \in O$. Hence, the resulting \tpmc $\pmc_\pomdp$ symbolically represents the POMDP $\pomdp$, and the direct \textsc{etr}-encoding in \zthree does not contain any $y$ variables. 
\Cref{figure:c-simplified} shows how the location \tpmc $\pmc_\mdp$ in \Cref{figure:c} is simplified to $\pmc_{\pomdp}$.
Additionally, our implementation supports algorithms using the oracle to assert further constraints to the solver, such as partial encodings of optimal strategies in the underlying MDPs $\mdp$. This feature simplifies the complexity of the \textsc{etr}-encoding for \zthree to a great extent, and enables a fraction of the parameters to be (virtually) instantiated.
 
\begin{figure}[h]
\vspace{-0.2cm}
\adjustbox{max width=\textwidth}{
	\begin{tikzpicture}[->,>=stealth',shorten >=1pt,auto,node distance=3cm,semithick]
  \tikzstyle{every state}=[fill=white,draw=black,text=black]

  \node[state,inner sep=2pt,minimum size=1pt]         (S0)                   {$s_0$};
  \node[state,inner sep=2pt,minimum size=1pt]         (S1) [right of=S0]     {$s_1$};
  \node[state,inner sep=2pt,minimum size=1pt] [fill=green!20]         (S2) [right of=S1]     {$s_2$};
  \node[state,inner sep=2pt,minimum size=1pt]         (S3) [right of=S2]     {$s_3$};
  \node[state,inner sep=2pt,minimum size=1pt]         (S4) [right of=S3]     {$s_4$};

  \path (S0) edge   [bend left=20]           node [above][align=right]{\footnotesize$1\cdot x_{0, r}$} (S1)
             edge   [loop above]             node [above][align=center] {\footnotesize$1\cdot x_{0, l} $} (S0)
        (S1) edge                            node[above][align=center] {\footnotesize$1\cdot x_{\bot,r}$} (S2)
             edge   [bend left=20]           node [below][align=center]{\footnotesize$1\cdot x_{\bot,l}$} (S0)
        (S2) edge   [loop above]             node [above]{} (S2)
        (S3) edge   [bend left=20]           node [above][align=center]  {\footnotesize$1\cdot x_{3, r}$} (S4)
             edge                            node [above][align=center]{\footnotesize$1\cdot x_{3,l}$} (S2)
        (S4) edge   [bend left=20]           node [below][align=center] {\footnotesize$1\cdot x_{\bot,l}$ } (S3)
             edge   [loop above]             node [above][align=center]{\footnotesize$1\cdot x_{\bot,r}$} (S4);
  \node[below=0.15cm of S0] (s1) {$@ s_0$};
  \node[below=0.15cm of S3] (s3) {$@ s_3$};
\end{tikzpicture}
}
\caption{Simplified location \tpmc $\pmc_\pomdp$ for the location POMDP $\pomdp = \mdpline\setObs{\obs}$ in \Cref{figure:c}, where $\obs(s_0) = @s_0, \obs(s_3)=@s_3, \obs(s_1) = \obs(s_4) = \bot$.}
\vspace{-0.4cm}
\label{figure:c-simplified}
\end{figure}

\subsection{PMC Tools as Oracles}\label{subsec:integration-storm}
Another orthogonal approach integrated \textit{probabilistic model checking (PMC)} tools as symbolic alternatives to the SMT-based oracle (cf. \Cref{subsec:smt-oracle}). Rather than computing feasible rewards under a threshold $\tau$ as in the SMT checker, these approaches consider approximations of optimal expected rewards for the POMDPs induced by given observation functions. To minimise the overhead of re-generating model files, we utilize pre-built and parametric PRISM models.

We assess two pathways for enforcing memory-less policies in PMC tools, based on the strategy determinism. For \textbf{deterministic and memory-less} strategies, we consider the inductive synthesis option implemented in the PAYNT solver \cite{inductive-synthesis-fsc-pomdp,paynt-fspp}. The tool combines abstraction-refinement (AR) and counterexamples to explore finite-state controllers for POMDPs, which are by default memory-less policies. We omit such integration with PAYNT, as we consider the SMT parameter synthesis performance to be satisfactory for practical dimensions and budgets in deterministic strategies.

In the case of \textbf{randomized and memory-less} policies, the suggested approach is to apply gradient-based optimization on the \textit{parametric Markov Chain (pMC)} corresponding to the input POMDP. Using the formulation in \cite{permissive-fsc-convert-pmc}, POMDPs are converted to parametric MCs such that the state space encodes an unitary memory bound. The transformation introduces probability parameters for each observation-action pair, with respect to which one can extract symbolic derivatives (\texttt{storm-pars}~\cite{storm_docs}). We then optimize these parameters with various Gradient Descent methods (\textit{momentum-based gradient descent}, \textit{Adam}\cite{adam-kingma-stoch-opt} and \textit{RMSprop}\cite{hinton-rmsprop}) as proposed in \cite{heck2021-gradient-descent-randomized-fsc}. Despite model sensitivity to rewards, few parameters yielded non-zero symbolic gradients, suggesting a non-convex or discontinuous parameter landscape in our investigations.

Due to unsatisfactory performance for memory-less strategies, we have employed Storm \cite{storm} through both the \texttt{storm-pomdp} utility and binders in the \texttt{stormpy} library to analyse proposed POMDPs in a \textbf{memory-full} setting. Both interfaces rely on the \textit{Belief Exploration} model checker, introduced in \cite{belief-exploration}, to under-approximate the expected reward in the queried POMDP $\mdp\setObs{\obs}$. The POMDP analysis in Storm is performed by determining a memory-bound strategy for a belief-MDP that tracks all belief evolutions within $\mdp\setObs{\obs}$.
Restricting the size of the memory allowed in POMDP controllers provides an upper-bound on rewards, as it intuitively leads to less accurate strategies:
\[
\MinExpRew{\mdp\setObs{\obs}} \leq \inf_{\sigma \in \OStrat_{fin(k)}} \ExpRew{\mdp\setObs{\obs}}{\sigma} \leq \MinExpRewP{\mdp\setObs{\obs}}\,
\]
where $\OStrat_{\mathit{fin}(k)}$ represents the set of strategies that may be defined with a finite-state machine of size $k$, such that the set of memory-less strategies $\StratP = \OStrat_{\mathit{fin}(1)}$. Note that the \textit{finite-state controllers} (FSC) need not necessarily be memory-less, as we have considered throughout most of the paper.
Since memory-less controllers are less performant, the key idea of our attempt is to utilise the FSC with bounded memory, synthesised by Storm, as a lower-bound (minimal) guarantee on the expected reward of positional strategies.

\section{Heuristics for Decomposition}\label{sec:informed-search-algorithms}

Using the POMDP oracles in \Cref{sec:integration-pomdp-oracles}, we performed an investigation (cf. \Cref{app:ssp-line-experiments}) to find the observation function with optimal expected reward for varying \SSP \Line instances. We observed that certain patterns emerge when instance parameters, such as dimensions, budgets and thresholds, are slightly modified, but they are insufficient for a greedy search algorithm. We develop a new paradigm for solving \POP/\SSP-instances via decomposition to POMDPs, using algorithms aimed at selecting a (good) observation function wisely from the vast space of plausible candidates.

\Cref{subsec:atomic-distinguishability-groups} establishes the necessary algorithm primers with equivalence relations on the underlying MDP of \POP instances. Then, we exploit such concepts in \Cref{subsec:informed-search-pop} to construct an offline heuristic-based enumeration algorithm over a sensible subset of candidate POMDPs. We show that it is sound, and that it is complete and efficient given sufficient budget, but incomplete in the general case.

\subsection{Atomic Distinguishability Groups}\label{subsec:atomic-distinguishability-groups}
Consider an observation \tpmc $\pmc_{\mdp}$ for a \POP-instance $(\mdp,B,\tau)$ and the set of optimal positional deterministic strategies $\StratPD^{*}(\mdp) \subseteq \StratPD(\mdp)$. We define the set of local optimal actions for a state $s$ as
\[
\alpha^{*}(s)
= \left \{a \in Act_\mdp \mid \exists \sigma^{*} \in \StratPD^{*}(\mdp), \sigma^{*}(a \mid s) > 0 \right\},
\]
and interchangeably refer to it as its \emph{optimal signature}. We can further define equivalence relations between states in $S_\mdp$ using this concept.

\begin{definition}[Strong action equivalence]
\label{def:strong-action-equivalence}
    Two states $s$ and $s'$ are \emph{strong action-equivalent} $s \sim_{R_{\sigma^*}} s'$ iff they have identical optimal signatures, \ie, $\alpha^{*}(s) = \alpha^{*}(s')$.
\end{definition}

The strong equivalence relation induces a partition of $S_\mdp$. We refer to equivalence classes $\llbracket \cdot \rrbracket_{}$ under the strong action equivalence as \textit{atomic distinguishability groups}, and denote by $\mathfrak{G}(\mdp)$ the quotient set of $S_\mdp$ induced by \Cref{def:strong-action-equivalence}, \ie $S_M / R_{\sigma^*} = \{\llbracket s \rrbracket_{R_{\sigma^*}} \mid s \in S_M\}$.

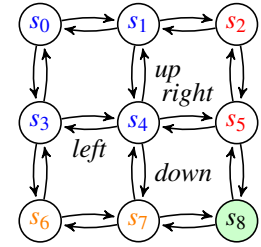
\begin{wrapfigure}{r}{0.28\linewidth}
\vspace{-0.9cm}
\centering
\begin{tikzpicture}[->,>=stealth',shorten >=1pt,auto,node distance=1.3cm,
                    semithick]
  \tikzstyle{every state}=[fill=white,draw=black,text=black]

  \node[state,inner sep=2pt,minimum size=1pt]         (S0)                   {\color{blue}{$s_0$}};
  \node[state,inner sep=2pt,minimum size=1pt]         (S1) [below of=S0]     {\color{blue}{$s_3$}};
  \node[state,inner sep=2pt,minimum size=1pt]         (S2) [below of=S1]     {\color{orange}{$s_6$}};
  \node[state,inner sep=2pt,minimum size=1pt]         (S3) [right of=S0]     {\color{blue}{$s_1$}};
  \node[state,inner sep=2pt,minimum size=1pt]         (S4) [right of=S1]     {\color{blue}{$s_4$}};
  \node[state,inner sep=2pt,minimum size=1pt]         (S5) [right of=S2]     {\color{orange}{$s_7$}};
  \node[state,inner sep=2pt,minimum size=1pt]         (S6) [right of=S3]     {\color{red}{$s_2$}};
  \node[state,inner sep=2pt,minimum size=1pt]         (S7) [right of=S4]     {\color{red}{$s_5$}};
  \node[state,inner sep=2pt,minimum size=1pt][fill=green!20]  (S8) [right of=S5]     {$s_8$};

  \path (S0) edge [bend left=10]             node {} (S3)
             edge  [bend right=10]             node [left]{} (S1)
        (S1) edge   [bend left=10]            node {} (S4)
             edge  [bend right=10]             node [above left] {} (S2)
             edge  [bend right=10]             node [right] {} (S0)
        (S2) edge     [bend left=10]          node {} (S5)
             edge    [bend right=10]           node [below right]{} (S1)
        (S3) edge    [bend right=10]           node [above left]{} (S4)
             edge   [bend left=10]            node {} (S6)
             edge     [bend left=10]          node {} (S0)
        (S4) edge    [bend left=10]           node {$\mathit{right}$} (S7)
             edge   [bend left=10]            node {$\mathit{down}$} (S5)
             edge   [bend left=10]            node {$\mathit{left}$} (S1)
             edge    [bend right=10]           node [ right] {$\mathit{up}$} (S3)
        (S5) edge   [bend left=10]            node {} (S4)
             edge   [bend left=10]            node {} (S8)
             edge     [bend left=10]          node {} (S2)
        (S6) edge    [bend left=10]           node {} (S7)
             edge    [bend left=10]           node {} (S3)
        (S7) edge     [bend left=10]          node {} (S4)
             edge     [bend left=10]          node {} (S8)
             edge    [bend left=10]           node {} (S6)
        (S8) edge    [bend left=10]           node  [above left]{} (S7)
             edge    [bend left=10]           node {} (S5);
\end{tikzpicture}
\caption{3x3 grid world $\mdp$ and $\mathfrak{G}(\mdp)$.} 
\vspace{-0.7cm}
\label{grid1-colored}
\end{wrapfigure}

\begin{example}\label{ex:atomic-distinguishability}
As an example, consider the $3\times3$ \Grid world in \Cref{grid1}, and let $\mdp$ be its underlying MDP.
We have the atomic distinguishability groups $\mathfrak{G}(\mdp) = \left \{ {\color{blue}{o_1}}, {\color{red}{o_2}}, {\color{orange}{o_3}} \right \}$, where ${\color{blue}{o_1}} = \left \{ s_0, s_1, s_3, s_4 \right \}$, ${\color{red}{o_2}} = \left \{ s_2, s_5 \right \}$, and ${\color{orange}{o_3}} = \left \{ s_6, s_7 \right \}$, as illustrated in \Cref{grid1-colored}. The optimal action set for the states in each group is $\left \{ \mathit{right}, \mathit{down} \right \}$, $\left \{ \mathit{down} \right \}$, and $\left \{ \mathit{right} \right \}$, respectively. 
Moreover, one can directly create an optimal observation function $\obs_{\mathfrak{G}}\colon S_\mdp \to \OO$ with $\lvert O \rvert = 3$ from these groups. 
\end{example}

Note that in \Cref{ex:atomic-distinguishability}, the minimal positional budget $B^{*}_\mdp$ (cf. \Cref{def:minimalBudget}) is 2. We can transform $\obs_{\mathfrak{G}}$ into an optimal observation function $\obs'\colon S_M \to \OOp$ with $\lvert O' \rvert = B^{*}_\mdp$ by combining ${\color{blue}{o_1}}$ and ${\color{red}{o_2}}$ into $o_{1,2} = o_1 \cup o_2$.

An optimal positional observation-based strategy for the POMDP $\mdp\setObs{\obs'}$ assigns $\mathit{down}$ for $o_{1,2}$ and $\mathit{right}$ for $o_{3}$. Informally, the transformed observation function $\obs'$ remains optimal since the states in the new observation class $o_{1,2}$ share a common optimal action, $\mathit{down}$. This introduces a weaker notion of the action equivalence relation that induces multiple partitions of $S_\mdp$.

\begin{definition}[Weak action equivalence]
    Two states $s$ and $s'$ are \emph{weakly action-equivalent} $s \sim s'$ if and only if $\alpha^{*}(s) \cap \alpha^{*}(s') \neq \emptyset$.
\end{definition}

\subsection{Offline Heuristic Enumeration Algorithm for \POP-instances}\label{subsec:informed-search-pop}

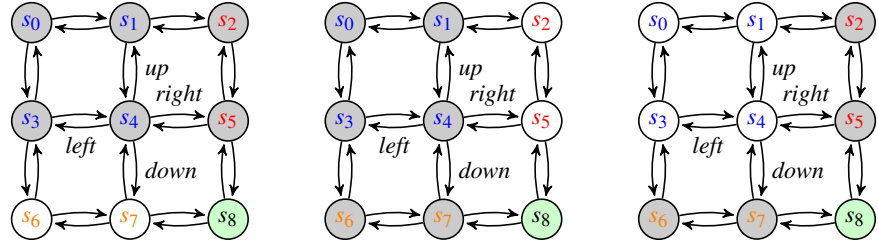
\begin{figure}[h]
\centering
%
\begin{minipage}{0.32\linewidth}
\centering
\begin{tikzpicture}[->,>=stealth',shorten >=1pt,auto,node distance=1.3cm,
                    semithick,scale=1,transform shape]
  \tikzstyle{every state}=[fill=white,draw=black,text=black]

  \node[state,inner sep=2pt,minimum size=1pt][fill=gray!40]         (S0)                   {\color{blue}{$s_0$}};
  \node[state,inner sep=2pt,minimum size=1pt][fill=gray!40]         (S1) [below of=S0]     {\color{blue}{$s_3$}};
  \node[state,inner sep=2pt,minimum size=1pt]                        (S2) [below of=S1]     {\color{orange}{$s_6$}};
  \node[state,inner sep=2pt,minimum size=1pt][fill=gray!40]         (S3) [right of=S0]     {\color{blue}{$s_1$}};
  \node[state,inner sep=2pt,minimum size=1pt][fill=gray!40]         (S4) [right of=S1]     {\color{blue}{$s_4$}};
  \node[state,inner sep=2pt,minimum size=1pt]                        (S5) [right of=S2]     {\color{orange}{$s_7$}};
  \node[state,inner sep=2pt,minimum size=1pt][fill=gray!40]         (S6) [right of=S3]     {\color{red}{$s_2$}};
  \node[state,inner sep=2pt,minimum size=1pt][fill=gray!40]         (S7) [right of=S4]     {\color{red}{$s_5$}};
  \node[state,inner sep=2pt,minimum size=1pt][fill=green!20]        (S8) [right of=S5]     {$s_8$};

  \path (S0) edge [bend left=10] (S3)
             edge [bend right=10] (S1)
        (S1) edge [bend left=10] (S4)
             edge [bend right=10] (S2)
             edge [bend right=10] (S0)
        (S2) edge [bend left=10] (S5)
             edge [bend right=10] (S1)
        (S3) edge [bend right=10] (S4)
             edge [bend left=10] (S6)
             edge [bend left=10] (S0)
        (S4) edge [bend left=10] node {$\mathit{right}$} (S7)
             edge [bend left=10] node {$\mathit{down}$} (S5)
             edge [bend left=10] node {$\mathit{left}$} (S1)
             edge [bend right=10] node [right] {$\mathit{up}$} (S3)
        (S5) edge [bend left=10] (S4)
             edge [bend left=10] (S8)
             edge [bend left=10] (S2)
        (S6) edge [bend left=10] (S7)
             edge [bend left=10] (S3)
        (S7) edge [bend left=10] (S4)
             edge [bend left=10] (S8)
             edge [bend left=10] (S6)
        (S8) edge [bend left=10] (S7)
             edge [bend left=10] (S5);
\end{tikzpicture}
\end{minipage}
\hfill
%
\begin{minipage}{0.32\linewidth}
\centering
\begin{tikzpicture}[->,>=stealth',shorten >=1pt,auto,node distance=1.3cm,
                    semithick,scale=1,transform shape]
  \tikzstyle{every state}=[fill=white,draw=black,text=black]

  \node[state,inner sep=2pt,minimum size=1pt][fill=gray!40]         (S0)                   {\color{blue}{$s_0$}};
  \node[state,inner sep=2pt,minimum size=1pt][fill=gray!40]         (S1) [below of=S0]     {\color{blue}{$s_3$}};
  \node[state,inner sep=2pt,minimum size=1pt][fill=gray!40]         (S2) [below of=S1]     {\color{orange}{$s_6$}};
  \node[state,inner sep=2pt,minimum size=1pt][fill=gray!40]         (S3) [right of=S0]     {\color{blue}{$s_1$}};
  \node[state,inner sep=2pt,minimum size=1pt][fill=gray!40]         (S4) [right of=S1]     {\color{blue}{$s_4$}};
  \node[state,inner sep=2pt,minimum size=1pt][fill=gray!40]         (S5) [right of=S2]     {\color{orange}{$s_7$}};
  \node[state,inner sep=2pt,minimum size=1pt]                        (S6) [right of=S3]     {\color{red}{$s_2$}};
  \node[state,inner sep=2pt,minimum size=1pt]                        (S7) [right of=S4]     {\color{red}{$s_5$}};
  \node[state,inner sep=2pt,minimum size=1pt][fill=green!20]        (S8) [right of=S5]     {$s_8$};

  \path (S0) edge [bend left=10] (S3)
             edge [bend right=10] (S1)
        (S1) edge [bend left=10] (S4)
             edge [bend right=10] (S2)
             edge [bend right=10] (S0)
        (S2) edge [bend left=10] (S5)
             edge [bend right=10] (S1)
        (S3) edge [bend right=10] (S4)
             edge [bend left=10] (S6)
             edge [bend left=10] (S0)
        (S4) edge [bend left=10] node {$\mathit{right}$} (S7)
             edge [bend left=10] node {$\mathit{down}$} (S5)
             edge [bend left=10] node {$\mathit{left}$} (S1)
             edge [bend right=10] node [right] {$\mathit{up}$} (S3)
        (S5) edge [bend left=10] (S4)
             edge [bend left=10] (S8)
             edge [bend left=10] (S2)
        (S6) edge [bend left=10] (S7)
             edge [bend left=10] (S3)
        (S7) edge [bend left=10] (S4)
             edge [bend left=10] (S8)
             edge [bend left=10] (S6)
        (S8) edge [bend left=10] (S7)
             edge [bend left=10] (S5);
\end{tikzpicture}
\end{minipage}
\hfill
%
\begin{minipage}{0.32\linewidth}
\centering
\begin{tikzpicture}[->,>=stealth',shorten >=1pt,auto,node distance=1.3cm,
                    semithick,scale=1,transform shape]
  \tikzstyle{every state}=[fill=white,draw=black,text=black]

  \node[state,inner sep=2pt,minimum size=1pt]                        (S0)                   {\color{blue}{$s_0$}};
  \node[state,inner sep=2pt,minimum size=1pt]                        (S1) [below of=S0]     {\color{blue}{$s_3$}};
  \node[state,inner sep=2pt,minimum size=1pt][fill=gray!40]         (S2) [below of=S1]     {\color{orange}{$s_6$}};
  \node[state,inner sep=2pt,minimum size=1pt]                        (S3) [right of=S0]     {\color{blue}{$s_1$}};
  \node[state,inner sep=2pt,minimum size=1pt]                        (S4) [right of=S1]     {\color{blue}{$s_4$}};
  \node[state,inner sep=2pt,minimum size=1pt][fill=gray!40]         (S5) [right of=S2]     {\color{orange}{$s_7$}};
  \node[state,inner sep=2pt,minimum size=1pt][fill=gray!40]         (S6) [right of=S3]     {\color{red}{$s_2$}};
  \node[state,inner sep=2pt,minimum size=1pt][fill=gray!40]         (S7) [right of=S4]     {\color{red}{$s_5$}};
  \node[state,inner sep=2pt,minimum size=1pt][fill=green!20]        (S8) [right of=S5]     {$s_8$};

  \path (S0) edge [bend left=10] (S3)
             edge [bend right=10] (S1)
        (S1) edge [bend left=10] (S4)
             edge [bend right=10] (S2)
             edge [bend right=10] (S0)
        (S2) edge [bend left=10] (S5)
             edge [bend right=10] (S1)
        (S3) edge [bend right=10] (S4)
             edge [bend left=10] (S6)
             edge [bend left=10] (S0)
        (S4) edge [bend left=10] node {$\mathit{right}$} (S7)
             edge [bend left=10] node {$\mathit{down}$} (S5)
             edge [bend left=10] node {$\mathit{left}$} (S1)
             edge [bend right=10] node [right] {$\mathit{up}$} (S3)
        (S5) edge [bend left=10] (S4)
             edge [bend left=10] (S8)
             edge [bend left=10] (S2)
        (S6) edge [bend left=10] (S7)
             edge [bend left=10] (S3)
        (S7) edge [bend left=10] (S4)
             edge [bend left=10] (S8)
             edge [bend left=10] (S6)
        (S8) edge [bend left=10] (S7)
             edge [bend left=10] (S5);
\end{tikzpicture}
\end{minipage}

\caption{Resulting 3 partitions for $S(3, 2)$ for the 3x3 grid world $\mdp$. States marked in {\color{gray}{\textbf{gray}}} belong to the same partition block.}
\label{grid-1-colored-mixed}
\vspace{-0.5cm}
\end{figure}

\smallskip
\noindent\textit{$A_\mathfrak{G}$ Algorithm.}
Let $S(n,k)$ denote the Stirling number of the second kind, \ie, the number of ways to partition a set of $n$ objects into $k$ non-empty indistinguishable sets.
First, we partition the set of $n = \lvert \mathfrak{G}(\mdp) \rvert $ atomic distinguishability group \textit{labels} into $k = \min (B, n)$ observation classes for a \POP-instance input $(\mdp,B,\tau)$, resulting in $S(n,k)$ partitions. 
Each partition $p = \left \{ b_{1}, \dots, b_{k} \right \}$, in which the blocks $b_{i} \subseteq \mathfrak{G}(\mdp)$, represents an observation function $\obs\colon S_\mdp \to \OO$ with $\lvert O \rvert = k$ such that $\obs(s) = i$ iff $s \in b_i$.
\Cref{ex:atomic-distinguishability} corresponds to a case with $S(3,3) = 1$. Trivially, the single partition of labels in this case would be $p = \left\{\{{\color{blue}{o_1}}\}, \{{\color{red}{o_2}}\}, \{{\color{orange}{o_3}}\} \right\}.$ For $S(3,2) = 3$, \Cref{grid-1-colored-mixed} illustrates the possible partitions $\left\{\{{\color{blue}{o_1}}, {\color{red}{o_2}}\}, \{{\color{orange}{o_3}}\} \right\}$, $\left\{\{{\color{blue}{o_1}}, {\color{orange}{o_3}}\}, \{{\color{red}{o_2}}\} \right\}$, and $\left\{\{{\color{blue}{o_1}}\}, \{{\color{red}{o_2}}, {\color{orange}{o_3}}\} \right\}$.

%
%
Then, each partition (and its induced observation function) is ranked decreasingly by (1) its equivalence score, and (2) the number of strategy constraints that can be inferred about it. For clarity, the equivalence score $\mathit{eq}(\obs)$ of an observation function $\obs\colon S_\mdp \to \OO$ is the number of observations for which all states assigned to it have a weak action equivalence. Formally, it is defined as
\vspace{-0.16cm}
\[
    \mathit{eq}(\obs) = \sum_{o \in O} \mathit{eq}(o), ~\text{where}~ \mathit{eq}(o)=~ 
    \begin{cases}
      1, & ~\text{if}~ \bigcap_{s \in \obs^{-1}(o)} \alpha^{*}(s) \neq \emptyset \\
    0, & ~\text{otherwise.}
   \end{cases}
\vspace{-0.16cm}
\]

Finally, for every partition, the POMDP generated by the encoded observation function is evaluated using the SMT oracle introduced in \Cref{subsec:smt-oracle} together with the inferred strategy constraints asserted to the solver. If a (\texttt{sat}) solution is found, the algorithm returns it; otherwise, it returns \texttt{unknown}. Early termination occurs if $k\geq B^{*}_{\mdp}$ and the first POMDP evaluation returns \texttt{unsat}.

\smallskip
\noindent\textit{Heuristics.}
The heuristic components (1) and (2) are related. Let us consider a candidate observation function $\obs\colon S_\mdp \to \OO$ with $\lvert O \rvert \leq B$ for the \POP-instance $(\mdp,B,\tau)$, which is encoded by the \tpmc $\pmc_\mdp$. For each observation class $o \in O$, if $eq(o) = 1$, one can fix the value of the $x_{o,a}$ variables in $V_{\pmc_\mdp}$ such that $\sum_{a \in \Act_{\mdp}(o)} x_{o,a} = 1$ and $\sum_{a \in \Act_\mdp \setminus \Act_\mdp(o)} x_{o,a} = 0$, where $\Act_\mdp(o) = \bigcap_{s \in \obs^{-1}(o)} \alpha^{*}(s)$. Otherwise, one can fix $x_{o,a} = 0$ for all  $a \notin \bigcup_{s \in obs^{-1}(o)} \alpha^{*}(s)$.
Therefore, the heuristic components prioritise observation functions that are both more likely to admit optimal actions across observation classes and faster to evaluate, ensuring that more constrained (and promising) POMDPs are examined first. In fact, if there exists an optimal observation function $\obs'\colon S_\mdp \to \OOp$ with $\lvert O' \rvert \leq B$ over positional strategies, \ie, $\MinExpRewP{\mdp\setObs{\obs'}} = \MinExpRew{\mdp}$, then such a function is the first to be synthesised by $A_\mathfrak{G}$. To see this, note that if $\MinExpRewP{\mdp\setObs{\obs'}} = \MinExpRew{\mdp}$, there exists an optimal positional observation-based strategy $\sigma$ such that $\ExpRew{\mdp\setObs{\obs'}}{\sigma} = \MinExpRew{\mdp}$. Since only optimal actions are selected for every observation class $o \in O'$, the set of actions selected $\left \{ a \in \Act_\mdp \mid \sigma(a \mid s) > 0 \right \}$ is a subset of $\alpha^{*}(s)$ and is identical for every $s \in \obs'^{-1}(o)$. In essence, all states in $o$ share a set of optimal actions with each other, \ie, the set of actions selected, and the heuristic components will be maximal for $o$. Since this holds for all $o \in O'$, only the ranking of optimal observation functions will be maximal, and therefore one of them will be the first to be evaluated.
Additionally, the heuristics in $A_\mathfrak{G}$ can be used to construct meaningful initializations for future research on reinforcement learning and informed search algorithms solving both \POP- and \SSP-instances \cite{msc-thesis}. We cover in detail how to apply the inherently \POP-based heuristics to make initial guesses for \SSP-instances in \Cref{app:mpbp-initial-guess}.

\smallskip
\noindent\textit{Correctness.}
Given a \POP-instance $(\mdp,B,\tau)$, the algorithm performs one oracle call for each candidate of the \emph{finite} subset of possible observation functions, and therefore terminates. Moreover, if the oracle returns \texttt{sat} for $\MinExpRewP{\pomdp} \leq \tau$, where $\pomdp = (M, O, \obs)$, we have a valid solution to the \POP-instance (since $\obs\colon S_\mdp \to \OO$ and $\lvert O \rvert \leq B$), and we return it. The early termination is also correct following \Cref{def:minimalBudget} and the property shown above for the heuristic components. Therefore, the algorithm is also sound. However, it only considers observation functions that join atomic distinguishability groups to form observation classes, \ie, that never split an atomic distinguishability group into multiple observation classes.
We present in \Cref{app:incompleteness-clustering} a case that shows that $A_\mathfrak{G}$ is not complete in the general case for budget $B \leq B^{*}$. It remains an open conjecture whether the algorithm is complete (and therefore correct) for the \Line, \Maze, and \Grid instances studied for such budgets. Intuitively, completeness requires that, for any solvable \POP-instance in these topologies, there always exists a valid observation function $\obs'$ that respects the weak action equivalence relation among non-goal states when assigning them an observation, \ie, $s \sim s' \implies \obs'(s) = \obs'(s')$.

\smallskip
\noindent\textit{Performance.}
We analyze the running time of $A_\mathfrak{G}$ on a \POP-instance $(\mdp,B,\tau)$ based on the number of POMDPs evaluated.
There may be one atomic distinguishability group per non-goal state and at most one for every set in the powerset of $\Act_\mdp$, so $A_\mathfrak{G}$ explores at most $S(n, k)$ observation functions, where $n = \min \left( \lvert S_\mdp \setminus G_\mdp \rvert, 2^{|\Act_\mdp|} \right)$ and $k = \min \left ( B, \lvert \mathfrak{G}(\mdp) \rvert \right )$.
Therefore, it explores a number of (promising) candidate solutions that is asymptotically smaller than the total number of all possible POMDPs, $B^{\lvert S_\mdp \setminus G_\mdp \rvert}$.

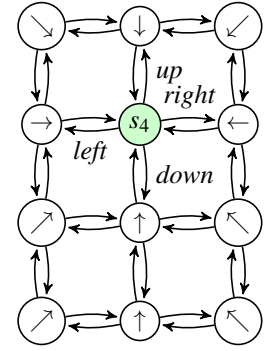
\begin{wrapfigure}{r}{0.28\textwidth}
\vspace{-0.85cm}
\begin{tikzpicture}[->,>=stealth',shorten >=1pt,auto,node distance=1.3cm,
                    semithick]
  \tikzstyle{every state}=[fill=white,draw=black,text=black]

  \node[state,inner sep=2pt,minimum size=1pt]         (S0) {\footnotesize$\searrow$};
  \node[state,inner sep=2pt,minimum size=1pt]         (S1) [below of=S0]     {\footnotesize$\rightarrow$};
  \node[state,inner sep=2pt,minimum size=1pt]         (S2) [below of=S1]     {\footnotesize$\nearrow$};
  \node[state,inner sep=2pt,minimum size=1pt]         (S3) [right of=S0]     {\footnotesize$\downarrow$};
  \node[state,inner sep=2pt,minimum size=1pt][fill=green!20] (S4) [right of=S1] {$s_4$};
  \node[state,inner sep=2pt,minimum size=1pt]         (S5) [right of=S2]     {\footnotesize$\uparrow$};
  \node[state,inner sep=2pt,minimum size=1pt]         (S6) [right of=S3]     {\footnotesize$\swarrow$};
  \node[state,inner sep=2pt,minimum size=1pt]         (S7) [right of=S4]     {\footnotesize$\leftarrow$};
  \node[state,inner sep=2pt,minimum size=1pt]         (S8) [right of=S5]     {\footnotesize$\nwarrow$};

  \node[state,inner sep=2pt,minimum size=1pt]         (S9) [below of=S2]     {\footnotesize$\nearrow$};
  \node[state,inner sep=2pt,minimum size=1pt]         (S10) [below of=S5]     {\footnotesize$\uparrow$};
  \node[state,inner sep=2pt,minimum size=1pt]         (S11) [below of=S8]     {\footnotesize$\nwarrow$};

  \path (S0) edge [bend left=10]             node {} (S3)
             edge  [bend right=10]             node [left]{} (S1)
        (S1) edge   [bend left=10]            node {} (S4)
             edge  [bend right=10]             node [above left] {} (S2)
             edge  [bend right=10]             node [right] {} (S0)
        (S2) edge     [bend left=10]          node {} (S5)
             edge    [bend right=10]           node [below right]{} (S1)
        (S3) edge    [bend right=10]           node [above left]{} (S4)
             edge   [bend left=10]            node {} (S6)
             edge     [bend left=10]          node {} (S0)
        (S4) edge    [bend left=10]           node {$\mathit{right}$} (S7)
             edge   [bend left=10]            node {$\mathit{down}$} (S5)
             edge   [bend left=10]            node {$\mathit{left}$} (S1)
             edge    [bend right=10]           node [ right] {$\mathit{up}$} (S3)
        (S5) edge   [bend left=10]            node {} (S4)
        edge   [bend left=10]            node {} (S8)
             edge     [bend left=10]          node {} (S2)
        (S6) edge    [bend left=10]           node {} (S7)
             edge    [bend left=10]           node {} (S3)
        (S7) edge     [bend left=10]          node {} (S4)
             edge     [bend left=10]          node {} (S8)
             edge    [bend left=10]           node {} (S6)
        (S8) edge    [bend left=10]           node  [above left]{} (S7)
             edge    [bend left=10]           node {} (S5)

        (S2) edge    [bend left=10]           node {} (S9)
        (S5) edge    [bend left=10]           node {} (S10)
        (S8) edge    [bend left=10]           node {} (S11)
        (S9) edge   [bend left=10]            node {} (S2)
             edge   [bend left=10]            node {} (S10)
        (S10) edge    [bend left=10]           node {} (S5)
             edge    [bend left=10]           node {} (S9)
             edge   [bend left=10]            node {} (S11)
        (S11) edge     [bend left=10]          node {} (S8)
             edge    [bend left=10]           node {} (S10);
\end{tikzpicture}
\caption{4x3 grid.} 
\vspace{-0.7cm}
\label{fig:grid4x3}
\end{wrapfigure}
Although $S(n,k)$ grows super-exponentially in $n$, we are limited to 966 POMDP evaluations in the (\Line, \Maze, \Grid) instances benchmarked. \Line instances only have 2 actions $\left \{ \mathit{left}, \mathit{right} \right \}$ and there is a single optimal action per state in the encoding MDP; it follows that $S(2, k) = 1$. Similarly for \Maze instances, which have 4 actions. Then, $S(4, k) \leq 7$. \Cref{fig:grid4x3} shows a \Grid instance where we indicate non-goal states by their atomic distinguishability group label. For instance, states belonging to group `` \scalebox{0.8}{$\nearrow$} '' have the optimal signature $\left \{ \mathit{up}, \mathit{right} \right \}$. Observe that the maximal number of atomic distinguishability groups for such instances is 8, as in the \Grid world studied, the set of optimal actions for a state does not contain actions in opposing directions (such as $\mathit{up}$ and $\mathit{down}$). Given there are 4 possible actions in this topology, $B^{*}_{\mdp} \leq 4$; moreover, we have shown that if $B \geq B^{*}_{\mdp}$, a single POMDP is evaluated. Hence, $S(8, k) = 1$ for $4 \leq k \leq 8$, and $S(8, k) \leq 966$ for $1 \leq k < 4$.

We validated the performance of the algorithm $A_\mathfrak{G}$ on the benchmark in \Cref{sec:enhancements-smt}, as shown in \Cref{tab:clustering-vs-sec3-time}. The underlying oracle for the $A_\mathfrak{G}$ algorithm in these results is the adapted SMT parameter synthesis approach, presented in \Cref{subsec:smt-oracle}.
We can see that decomposition to POMDPs with $A_\mathfrak{G}$ outperformed the solving time of our enhancements in \Cref{sec:enhancements-smt} by $2-3$ orders of magnitude.
It also scaled the dimensions of the solvable \Line, \Maze, and \Grid instances by factors greater than $100$, as shown in \Cref{tab:clustering-vs-sec3-size}.
To conclude, we evaluated its performance for \POP-instances with budget strictly less than their \textbf{MPB} (cf. \Cref{tab:clustering-vs-sec3-size}, right). As expected, the scale of solvable instances is significantly reduced because fewer constraints can be inferred on the POMDPs evaluated. However, we were still able to solve a new category of \POP-instances which was not previously solvable with our efforts in \Cref{sec:enhancements-smt} or the work in \cite{cav-base-oop}.

\begin{table}[ht!]
\vspace{-0.2cm}
\centering
\scriptsize
\begin{minipage}[t]{0.49\textwidth}
\begin{tabular}[t]{|l|l|l|C{1.1cm}|C{1.85cm}|} 
 \multicolumn{5}{c}{\textbf{PDOOP} - Deterministic Strategies} \\
 \hline
 \multicolumn{3}{|c|}{Problem Instance} & \Cref{sec:enhancements-smt} & $A_\mathfrak{G}$ (\Cref{subsec:informed-search-pop}) \\
 \hline
 Model & Thresh. & Budget & Time(s) & Time(s) \\
 \hline \hline


  \multirow{3}{*}{L$(377)$} &  $\leq 189$  & 2  & $0.0368$  & \improve{$0.004527$} \\ \cline{2-5}
  &  $\leq \frac{189}{2}$  & 2  & $0.0367$  & \improve{$0.004746$}  \\ \cline{2-5}
  &   $< \frac{189}{2}$    & 2  & $0.0344$ & \improve{$0.000172$} \\   

  \hline\hline

  \multirow{3}{*}{G$(24)$} &  $\leq \frac{26496}{575}$ & 2  & $0.0824$  & \improve{$0.012587$} \\ \cline{2-5}
  &  $\leq \frac{13248}{575}$ & 2  & $0.0838$  & \improve{$0.012067$} \\ \cline{2-5}
  &  $< \frac{13248}{575}$    & 2  & $0.0764$  & \improve{$0.000180$} \\  \hline\hline

  \multirow{3}{*}{M$(39)$} &  $\leq \frac{6232}{95}$  & 4  & $0.0265$ & \improve{$0.001722$} \\ \cline{2-5}
  &  $\leq \frac{3116}{95}$  & 4  & $0.0262$ & \improve{$0.001795$} \\ \cline{2-5}
  &  $< \frac{3116}{95}$     & 4  & $0.0227$ & \improve{$0.000157$} \\ 
 \hline
\end{tabular}
\end{minipage}
\hfill
\begin{minipage}[t]{0.49\textwidth}
\begin{tabular}[t]{|l|l|l|C{1.1cm}|C{1.85cm}|} 
 \multicolumn{5}{c}{\POP - Randomized Strategies} \\
 \hline
 \multicolumn{3}{|c|}{Problem Instance} & \Cref{sec:enhancements-smt} & $A_\mathfrak{G}$ (\Cref{subsec:informed-search-pop}) \\
 \hline
 Model & Thresh. & Budget & Time(s) & Time(s) \\
 \hline \hline

  \multirow{3}{*}{L$(249)$} &  $\leq \frac{250}{2}$  & 2  & \timeout{} & \improve{$0.003467$} \\ \cline{2-5}
  & $ \leq \frac{125}{2}$ & 2 & $0.2715$ & \improve{$0.003677$} \\ \cline{2-5}
  & $< \frac{125}{2}$ & 2 & $0.0254$ & \improve{$0.000152$} \\ 

  \hline\hline

  \multirow{3}{*}{G$(20)$} &  $\leq \frac{15200}{399}$ & 2  & \timeout{} & \improve{$0.006316$}  \\ \cline{2-5}
  &  $\leq \frac{7600}{399}$ & 2  & $30.896$  & \improve{$0.006041$} \\ \cline{2-5}
  &  $< \frac{7600}{399}$    & 2  & $0.0502$  & \improve{$0.000162$} \\ 
  
  \hline\hline
  
  \multirow{3}{*}{M$(7)$} &  $\leq \frac{168}{15}$  & 4  & \timeout{}  & \improve{$0.000894$} \\ \cline{2-5}
  &  $\leq \frac{84}{15}$  & 4  & $0.1849$  & \improve{$0.000802$} \\ \cline{2-5}
  &  $< \frac{84}{15}$     & 4  & $0.0082$ & \improve{$0.000202$} \\ 
 \hline

\end{tabular}
\end{minipage}
\vspace{0.1cm}
\caption{Runtime performance of $A_\mathfrak{G}$ via decomposition to POMDPs compared to our enhancements in \Cref{sec:enhancements-smt}.}
\label{tab:clustering-vs-sec3-time}
\vspace{-0.2cm}
\end{table}

\begin{table}[ht!]
\vspace{-0.2cm}
\centering
\scriptsize
\begin{minipage}[t]{0.31\textwidth}
\begin{tabular}[t]{|l|c|c|} 
 \multicolumn{3}{c}{\textbf{PDOOP} - Deterministic Strategies} \\
 \hline
 \multirow{2}{*}{World} & \Cref{sec:enhancements-smt} & $A_\mathfrak{G}$ \\ \cline{2-3}
  & $\lvert S_M \rvert$ & $\lvert S_M \rvert$ \\ \hline \hline

  \Line & $10~001$ & \improve{$200~001$} \\ \hline
  \Grid & $1~296$ & \improve{$160~801$} \\ \hline
  \Maze & $846$ & \improve{$200~011$} \\ \hline
\end{tabular}
\end{minipage}
\hfill
\begin{minipage}[t]{0.29\textwidth}
\begin{tabular}[t]{|c|c|c|} 
 \multicolumn{3}{c}{\POP\ - Rand. Strategies ($B=B^{*}$)} \\
 \hline
 \multirow{2}{*}{World} & \Cref{sec:enhancements-smt} & $A_\mathfrak{G}$ \\ \cline{2-3}
  & $\lvert S_M \rvert$ & $\lvert S_M \rvert$ \\ \hline \hline

  \Line & $1~001$ & \improve{$100~001$} \\ \hline
  \Grid & $400$ & \improve{$100~489$} \\ \hline
  \Maze & $266$ & \improve{$100~001$} \\ \hline
\end{tabular}
\end{minipage}
\hfill
\begin{minipage}[t]{0.38\textwidth}
\hfill
\begin{tabular}[t]{|l|l|c|c|c|} 
 \multicolumn{5}{c}{\POP\ - Randomized Strategies ($B = B^{*}-1$)} \\
 \hline
 \multicolumn{3}{|c|}{Problem Instance} & \Cref{sec:enhancements-smt} & $A_\mathfrak{G}$ \\
 \hline
 Model & Thresh. & Budget & Time(s) & Time(s) \\
 \hline \hline
 
  G$(7)$  & $\leq 29$ & 3 & \timeout{} & \improve{$0.006316$} \\ \hline
  M$(25)$ & $\leq 31$ & 3 & \timeout{} & \improve{$0.000894$} \\ \hline

\end{tabular}
\end{minipage}
\vspace{0.1cm}
\caption{Largest solvable instances by $A_\mathfrak{G}$ via decomposition to POMDPs compared to our results in \Cref{sec:enhancements-smt}, measured by the number of states. For all instances in the tables, the goal is the centermost state.}
\label{tab:clustering-vs-sec3-size}
\vspace{-0.5cm}
\end{table}

\section{Conclusion \& Discussion}\label{sec:conclusion-future-work}

In this work, we have focused on combining symbolic and sub-symbolic AI techniques to scale planning for decidable variants of the Optimal Observability Problem (\OOP). More specifically, we exploited the potential of the SMT-based parameter synthesis introduced in \cite{cav-base-oop} for general and well-defined tpMCs. This approach improved solving time by a factor of $\sim 1000$ and solvable instance size by $\sim 75$ times. Next, we implemented oracles that evaluate POMDP rewards through both an adaptation of the SMT encoding and an integration of Storm controller synthesis. This enabled a new paradigm for solving \POP/\SSP instances via decomposition, where one explores the space of plausible observation functions.
We developed a heuristic-based enumeration algorithm following this paradigm that, compared to our own enhancements of the SMT-based parameter synthesis approach, achieves speedups of factors of $10^3$ on common benchmarks, and scales to instances that 100 times larger.

For future work, we believe that predictability for SMT can be further improved with \zthree tactics by decomposing queries into a sequence of pre-processing transformations. Additionally, scalability can potentially improve by extending parameter synthesis tools and POMDP controller synthesis methods.
Furthermore, several Reinforcement Learning techniques can be applied at different parameter layers based on oracle calls and/or simulations. The heuristic components for decomposition can be used to construct an initialization for such algorithms.
\vspace{-0.2cm}

\section*{Data Availability Statement:}
\vspace{-.2cm}
The entire source code and repository of benchmarks at various stages of refinements are available as open-source in a \href{https://github.com/Zedrichu/Optimal-Observability-Problem/tree/main}{\color{blue}GitHub Repository}.
\vspace{-.3cm}

%
%
%
%
\bibliographystyle{splncs04}
\bibliography{bibliografy.bib} 

\newpage
\appendix
\renewcommand{\theHsection}{A\arabic{section}}
\renewcommand{\theHsubsection}{A\arabic{section}.\arabic{subsection}}
\crefalias{section}{appendix}
\crefalias{subsection}{appendix}

\clearpage
\section{Appendix: SMT Parameter Synthesis}

\subsection{SMT Constraint Experiments}\label{app:smt-constraint-experiments}
\begin{figure}
\vspace{-0.6cm}
\centering
\includegraphics[width=0.8\textwidth]{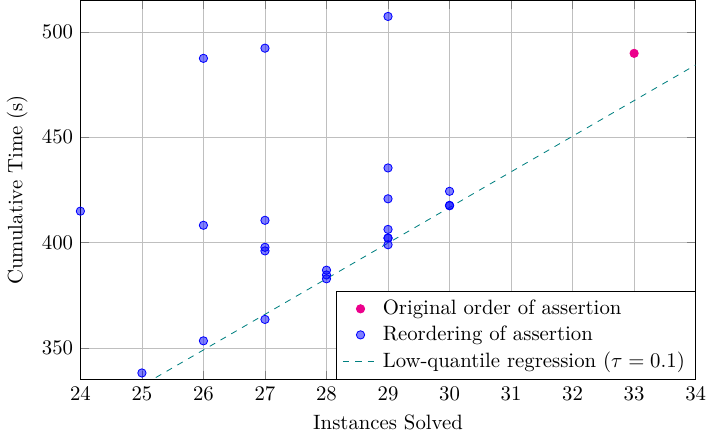}
\caption{Time to verify successful benchmark instances for various reorderings of constraint assertion.}
\label{fig:reorder-constraints}
\vspace{-0.6cm}
\end{figure}
The \textsc{x}-axis indicates the number of benchmark instances (from a total of 42) solved when applying each permutation. We measure on the \textsc{y}-axis the cumulative runtime for each reordering, \ie, the sum of runtimes for benchmark instances that did not time out.
 
The {\color{teal}teal} line indicates the low-quantile regression on the benchmark results of various reorderings, which are measured by their average solving time per (solvable) problem instance. Unsurprisingly, we observed that the solver's verification performance may be affected by over 25\%. The reorderings solved fewer instances, and many took a larger average solving time per instance compared to the original order of assertion.

\subsection{Native Boolean Encoding}\label{appendix:bool-enc}
Here we summarise the full tpMC encodings of each problem variant under Boolean encoding. The \emph{full observability} and \emph{threshold} constraints defined in \cite{msc-thesis} still apply to this encoding.
The typing constraints for observation assignments and deterministic strategies are reformulated as logical cardinalities:
\begin{align*}
    \forall_{o \in O}&\colon \bigvee\limits_{a \in Act} x_{o,a} \\
    \forall_{o \in O}&\colon \forall_{a \in Act}:
x_{o,a} \rightarrow \bigwedge\limits_{a' \in Act \setminus \{a\}} \neg x_{o, a'}
    &\\
    \forall_{s \in S \setminus G}&\colon\bigvee\limits_{o \in O} y_{s,o} \\
    \forall_{s \in S \setminus G}&\colon \forall_{o \in O}\colon y_{s,o} \rightarrow \bigwedge\limits_{o' \in O \setminus \{o\}} \neg y_{s, o'}
\end{align*}
Alternatively, to type observation assignment parameters as proper probability distributions in \POP, we can use Pseudo-boolean equality constraints to speed up the solver:
\begin{equation*}
    \forall s \in S\setminus G\colon \sum \limits_{o\in O} y_{s,o}*1 = 1
\end{equation*}

The system of Bellman equations is rewritten to accommodate native Boolean parameters:
\begin{center}
\hspace{-1cm}
\begin{tabular}{ c l } 
 Deterministic POP
 & $\forall_{s \in S \setminus G}\colon \forall_{o \in Obs}\colon \forall_{a \in Act}\colon y_{s,o} \wedge x_{o,a} \rightarrow v_{s} == 1 + \sum\limits_{t\in \mathit{Post}(s,a)} P_\mdp(s,a)(t)~v_t$ \\
 Deterministic SSP
 & $\forall_{s \in S \setminus G}\colon
\forall_{a \in Act}\colon
(y_{s} \land x_{s,a} \lor \neg y_{s} \land x_{\bot,a}) \rightarrow v_{s} == 1 + \sum\limits_{t\in \mathit{Post}(s,a)} P_\mdp(s,a)(t)~v_{t}$ \\  
 Randomised POP
 & $\forall_{s \in S \setminus G}\colon \forall_{o \in Obs}\colon y_{s,o} \rightarrow v_{s} == 1 + \sum\limits_{a \in Act} \sum\limits_{t\in \mathit{Post}(s,a)} P_\mdp(s,a)(t)~v_{t} \cdot x_{o,a}$ \\
 Randomised SSP
 & $\forall_{s \in S \setminus G}\colon 
y_{s} \rightarrow v_{s} == 1 + \sum \limits_{a \in Act}\sum\limits_{t\in \mathit{Post}(s,a)} P_\mdp(s,a)(t)~v_{t} \cdot x_{s,a} $ \\

 & $\forall_{s \in S \setminus G}\colon 
\neg y_{s} \rightarrow v_{s} == 1 + \sum \limits_{a \in Act}\sum\limits_{t\in \mathit{Post}(s,a)} P_\mdp(s,a)(t)~v_{t} \cdot x_{\bot,a}$ 
\end{tabular}
\end{center}

Finally, for \SSP the budget constraint is also implemented as a Pseudo-Boolean cardinality constraint:
\begin{equation*}
    \sum \limits_{s\in S\setminus G} y_s*1 == B
\end{equation*}
\clearpage
\section{Appendix: Maze example}\label{app:mazeExample}

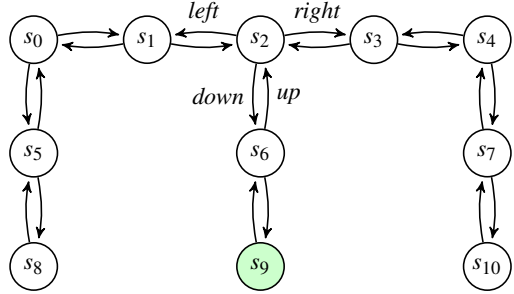
\begin{figure}[]
\centering
\begin{tikzpicture}[->,>=stealth',shorten >=1.5pt,auto,node distance=1.5cm,
                    semithick]
  \tikzstyle{every state}=[fill=white,draw=black,text=black]

  \node[state,inner sep=3pt,minimum size=1pt]         (S0)                   {$s_0$};
  \node[state,inner sep=3pt,minimum size=1pt]         (S1) [right of=S0]     {$s_1$};
  \node[state,inner sep=3pt,minimum size=1pt]         (S2) [right of=S1]     {$s_2$};
  \node[state,inner sep=3pt,minimum size=1pt]         (S3) [right of=S2]     {$s_3$};
  \node[state,inner sep=3pt,minimum size=1pt]         (S4) [right of=S3]     {$s_4$};
  \node[state,inner sep=3pt,minimum size=1pt]         (S5) [below of=S0]     {$s_5$};
  \node[state,inner sep=3pt,minimum size=1pt]         (S6) [below of=S2]     {$s_6$};
  \node[state,inner sep=3pt,minimum size=1.5pt]         (S7) [below of=S4]     {$s_7$};
  \node[state,inner sep=3pt,minimum size=1pt]         (S8) [below of=S5]     {$s_8$};
  \node[state,inner sep=3pt,minimum size=2.5pt][fill=green!20]          (S9) [below of=S6]     {$s_9$};
  \node[state,inner sep=2pt,minimum size=1pt]         (S10) [below of=S7]     {$s_{10}$};

  \path (S0) edge [bend left=10]             node {} (S1)
             edge  [bend right=10]           node {} (S5)
        (S1) edge   [bend right=10]          node {} (S2)
             edge  [bend left=10]            node {} (S0)
        (S2) edge     [bend right=10]        node [above ] {$\mathit{left}$} (S1)
             edge    [bend left=10]          node {$\mathit{right}$} (S3)
             edge    [bend right=10]         node [left]{$\mathit{down}$} (S6)
        (S3) edge    [bend right=10]         node{} (S4)
             edge   [bend left=10]           node {} (S2)
        (S4) edge    [bend right=10]         node {} (S3)
             edge   [bend left=10]           node {} (S7)
        (S5) edge   [bend right=10]          node {} (S0)
             edge   [bend left=10]           node {} (S8)
        (S6) edge    [bend right=10]         node [right] {$\mathit{up}$} (S2)
             edge    [bend left=10]          node {} (S9)
        (S7) edge     [bend left=10]         node {} (S4)
             edge     [bend left=10]         node {} (S10)
        (S8) edge    [bend left=10]          node {} (S5)
        (S9) edge    [bend left=10]          node {} (S6)
        (S10) edge    [bend left=10]         node {} (S7);
\end{tikzpicture}
\caption{Maze from~\cite{MCCALLUM1993190}.}
    \label{fig:mazegraph}
\end{figure}

\begin{example}[Agent in a Maze]
Another classical variant of our grid-like models is the maze depicted in  Figure~\ref{fig:mazegraph}, which was selected from ~\cite{MCCALLUM1993190}. An agent is placed in a random location on the maze and needs to find its goal by moving $\{\mathit{left, right, up, down}\}$. For simplicity, we omit the self-loops. The goal is at state $s_9$.
    The agent wants to reach the goal within the minimum number of steps. In the case of full observability, the agent is always able to distinguish the states and choose the optimal action. Given the underlying MDP $\mdp$ and an optimal positional deterministic strategy $\sigma\colon S_\mdp \rightarrow Act_\mdp$, where $\sigma(s_8) = \sigma(s_5) = \sigma(s_7) = \sigma(s_{10}) = up$, $\sigma(s_0) = \sigma(s_1) = right$, $\sigma(s_3) = \sigma(s_4) = left$, $\sigma(s_2) = \sigma(s_6) = down$.  We can construct a POMDP $\pomdp = (\mdp, O, obs)$, where $O = range(\sigma)$ and $obs(s) = \sigma(s)$. Hence, we need $|range(\sigma)| = 4$ observations and the following observation function: $obs(s_8) = obs(s_5) = obs(s_7) = obs(s_{10})$ and $obs(s_0) = obs(s_1)$ and $obs(s_3) = obs(s_4)$ and $obs(s_2) = obs(s_6)$. By choosing this observation function we get $\MinExpRewPD \mdp = \MinExpRewPD \pomdp$.
\end{example}

In our experiments, we consider instances of the maze that can only include an odd number of columns, but we are free to choose the number of rows. For the experiments, we increase the number of columns by $2$ each time, while the number of rows by $1$. For example for maze $M(5)$ as in~\Cref{fig:mazegraph} the number of rows is $3$. The next maze $M(7)$ consists of $7$ columns and $4$ rows. The number states of the maze can be calculated as $\#columns + 3\times(\#rows - 1)$. 

\clearpage
\section{Appendix: \SSP \Line experiments}\label{app:ssp-line-experiments}

\begin{figure}[h]{}
\centering
\begin{tikzpicture}[->,>=stealth',shorten >=1pt,auto,node distance=1.8cm, semithick]

  \tikzstyle{every state}=[fill=white,draw=black,text=black]

  \node[state,inner sep=2pt,minimum size=1pt]         (S0)                   {$s_0$};
  \node[state,inner sep=2pt,minimum size=1pt]         (S1) [right of=S0]     {$s_1$};
  \node[state,inner sep=2pt,minimum size=1pt]         (S2) [right of=S1] {$s_2$};
  \node[state,inner sep=2pt,minimum size=1pt][fill=green!35] (S3) [right of=S2] {$s_3$};
  \node[state,inner sep=2pt,minimum size=1pt]         (S4) [right of=S3]     {$s_4$};
  \node[state,inner sep=2pt,minimum size=1pt]         (S5) [right of=S4]     {$s_5$};
  \node[state,inner sep=2pt,minimum size=1pt]         (S6) [right of=S5]     {$s_6$};

  \node[below=0.15cm of S0][text=gray] (s0) {$@ s_0$};
  \node[below=0.15cm of S1][text=gray] (s1) {$@ s_1$};
  \node[below=0.15cm of S2][text=gray] (s2) {$@ s_2$};
  \node[below=0.15cm of S4][text=gray] (s4) {$@ s_4$};
  \node[below=0.15cm of S5][text=gray] (s5) {$@ s_5$};
  \node[below=0.15cm of S6][text=gray] (s6) {$@ s_6$};

  \path (S0) edge   [bend left=20]           node [above]{$r$} (S1)
             edge   [loop above]             node [right=3pt] {$\ell$} (S0)
        (S1) edge   [bend left=20]           node {$r$} (S2)
             edge   [bend left=20]           node [below]{$\ell$} (S0)
        (S2) edge                            node {$r$} (S3)
             edge   [bend left=20]           node [below]{$\ell$} (S1)
        (S3) edge   [loop above]             node [right=3pt]{$\ell$,$r$} (S3)
        (S4) edge                            node [above]{$\ell$} (S3)
             edge   [bend left=20]           node [above]{$r$} (S5)
        (S5) edge   [bend left=20]           node [above] {$r$} (S6)
             edge   [bend left=20]           node [below]{$\ell$} (S4)
        (S6) edge   [bend left=20]           node [below]{$\ell$} (S5)
             edge   [loop above]             node [left=3pt] {$r$} (S6);
\end{tikzpicture}
\caption{MDP $\mdpline$ with $p=1$ before sensors are selected (and an observation function is defined).}
\label{fig:mdpline-7}
\end{figure}
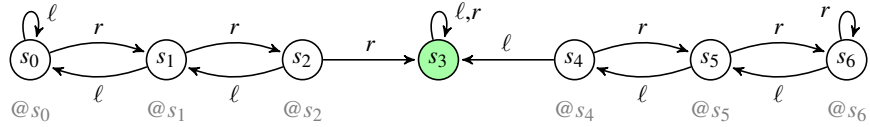

Consider some \SSP-instance $(\mdpline, B, \tau)$ for $\mdpline$ scaled to 7 states, as illustrated in \Cref{fig:mdpline-7}. For $B \geq 3$, it is trivial to see that either activating sensors $\{ @s_0, @s_1, @s_2 \}$ or $\{ @s_4, @s_5, @s_6 \}$ achieves optimal minimal expected reward of 2. However, uncertainty rises for a lower budget, \eg, $B=2$. Should one sensor be placed on each side of the goal, or should both be placed on the same side? If so, should they be side-by-side or spread away from each other? Can the expected reward be lower if the sensors are activated next to the goal or on the edge of the line? There are 9 non-symmetric observation functions for $B=2$. Which one(s), if any, would be feasible when $\tau = 5$, $\tau=4$, or even $\tau=3$?

In order to answer these questions, we made a series of brute-force approach experiments that challenged the feasibility of the observation functions for decreasing $\tau$. The results show that, for \Cref{fig:mdpline-7}, activating sensors $\left \{ @s_1, @s_2 \right\}$ or $\left \{ @s_5, @s_6 \right\}$ accounts for the lowest minimal expected reward of $\sim 3.216$, whereas $\left \{ @s_0, @s_6 \right\}$ has the largest minimal expected reward of $\frac{22}{4} \approx 7.333$.

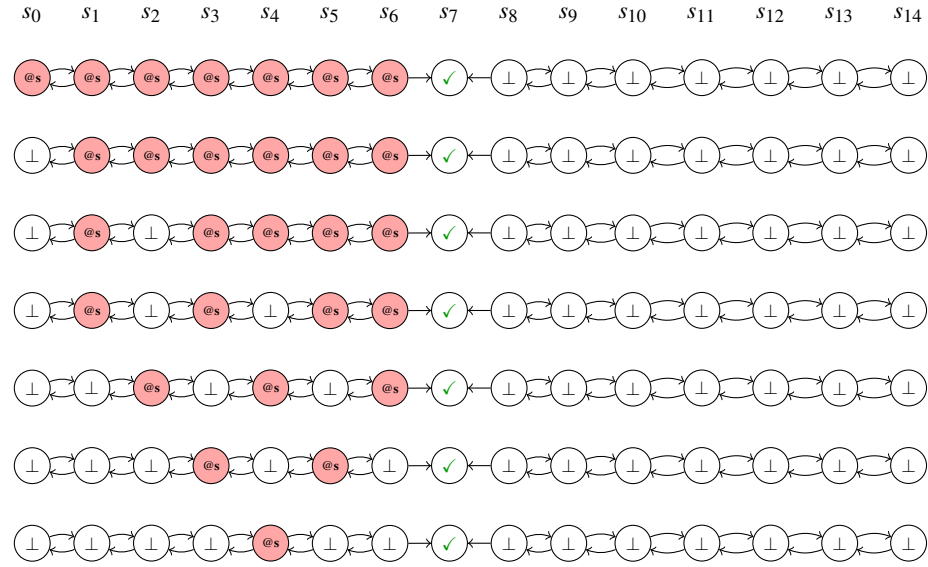
\begin{figure}[ht!]
\centering
\begin{tikzpicture}
    \tikzstyle{every state}=[fill=white,draw=black,text=black,inner sep=2pt,minimum size=1.5em]
    \tikzset{node distance=1.7em and 0.3cm}
    
    \begin{scope} [name prefix=label-, every node/.append style={draw=none}]
        \node (s0) [] {$s_0$};
        \node (s1) [right=of s0] {$s_1$};
        \node (s2) [right=of s1] {$s_2$};
        \node (s3) [right=of s2] {$s_3$};
        \node (s4) [right=of s3] {$s_4$};
        \node (s5) [right=of s4] {$s_5$};
        \node (s6) [right=of s5] {$s_6$};
        \node (s7) [right=of s6] {$s_7$};
        \node (s8) [right=of s7] {$s_8$};
        \node (s9) [right=of s8] {$s_9$};
        \node (s10) [right=of s9] {$s_{10}$};
        \node (s11) [right=of s10] {$s_{11}$};
        \node (s12) [right=of s11] {$s_{12}$};
        \node (s13) [right=of s12] {$s_{13}$};
        \node (s14) [right=of s13] {$s_{14}$};
    \end{scope}

    \begin{scope} [name prefix=budget7-, node distance=1.2em]
        \node[state,fill=red!35] (s0) [below=of label-s0] {\tiny$\mathbf{@s}$};
        \node[state,fill=red!35] (s1) [below=of label-s1] {\tiny$\mathbf{@s}$};
        \node[state,fill=red!35] (s2) [below=of label-s2] {\tiny$\mathbf{@s}$};
        \node[state,fill=red!35] (s3) [below=of label-s3] {\tiny$\mathbf{@s}$};
        \node[state,fill=red!35] (s4) [below=of label-s4] {\tiny$\mathbf{@s}$};
        \node[state,fill=red!35] (s5) [below=of label-s5] {\tiny$\mathbf{@s}$};
        \node[state,fill=red!35] (s6) [below=of label-s6] {\tiny$\mathbf{@s}$};
        \node[state] (s7) [below=of label-s7] {\obsGoalSS};
        \node[state] (s8) [below=of label-s8] {\scriptsize$\bot$};
        \node[state] (s9) [below=of label-s9] {\scriptsize$\bot$};
        \node[state] (s10) [below=of label-s10] {\scriptsize$\bot$};
        \node[state] (s11) [below=of label-s11] {\scriptsize$\bot$};
        \node[state] (s12) [below=of label-s12] {\scriptsize$\bot$};
        \node[state] (s13) [below=of label-s13] {\scriptsize$\bot$};
        \node[state] (s14) [below=of label-s14] {\scriptsize$\bot$};
    \end{scope}

    \begin{scope} [name prefix=budget6-]
        \node[state] (s0) [below=of budget7-s0] {\scriptsize$\bot$};
        \node[state,fill=red!35] (s1) [below=of budget7-s1] {\tiny$\mathbf{@s}$};
        \node[state,fill=red!35] (s2) [below=of budget7-s2] {\tiny$\mathbf{@s}$};
        \node[state,fill=red!35] (s3) [below=of budget7-s3] {\tiny$\mathbf{@s}$};
        \node[state,fill=red!35] (s4) [below=of budget7-s4] {\tiny$\mathbf{@s}$};
        \node[state,fill=red!35] (s5) [below=of budget7-s5] {\tiny$\mathbf{@s}$};
        \node[state,fill=red!35] (s6) [below=of budget7-s6] {\tiny$\mathbf{@s}$};
        \node[state] (s7) [below=of budget7-s7] {\obsGoalSS};
        \node[state] (s8) [below=of budget7-s8] {\scriptsize$\bot$};
        \node[state] (s9) [below=of budget7-s9] {\scriptsize$\bot$};
        \node[state] (s10) [below=of budget7-s10] {\scriptsize$\bot$};
        \node[state] (s11) [below=of budget7-s11] {\scriptsize$\bot$};
        \node[state] (s12) [below=of budget7-s12] {\scriptsize$\bot$};
        \node[state] (s13) [below=of budget7-s13] {\scriptsize$\bot$};
        \node[state] (s14) [below=of budget7-s14] {\scriptsize$\bot$};
    \end{scope}
    
    \begin{scope} [name prefix=budget5-]
        \node[state] (s0) [below=of budget6-s0] {\scriptsize$\bot$};
        \node[state,fill=red!35] (s1) [below=of budget6-s1] {\tiny$\mathbf{@s}$};
        \node[state] (s2) [below=of budget6-s2] {\scriptsize$\bot$};
        \node[state,fill=red!35] (s3) [below=of budget6-s3] {\tiny$\mathbf{@s}$};
        \node[state,fill=red!35] (s4) [below=of budget6-s4] {\tiny$\mathbf{@s}$};
        \node[state,fill=red!35] (s5) [below=of budget6-s5] {\tiny$\mathbf{@s}$};
        \node[state,fill=red!35] (s6) [below=of budget6-s6] {\tiny$\mathbf{@s}$};
        \node[state] (s7) [below=of budget6-s7] {\obsGoalSS};
        \node[state] (s8) [below=of budget6-s8] {\scriptsize$\bot$};
        \node[state] (s9) [below=of budget6-s9] {\scriptsize$\bot$};
        \node[state] (s10) [below=of budget6-s10] {\scriptsize$\bot$};
        \node[state] (s11) [below=of budget6-s11] {\scriptsize$\bot$};
        \node[state] (s12) [below=of budget6-s12] {\scriptsize$\bot$};
        \node[state] (s13) [below=of budget6-s13] {\scriptsize$\bot$};
        \node[state] (s14) [below=of budget6-s14] {\scriptsize$\bot$};
    \end{scope}

    \begin{scope} [name prefix=budget4-]
        \node[state] (s0) [below=of budget5-s0] {\scriptsize$\bot$};
        \node[state,fill=red!35] (s1) [below=of budget5-s1] {\tiny$\mathbf{@s}$};
        \node[state] (s2) [below=of budget5-s2] {\scriptsize$\bot$};
        \node[state,fill=red!35] (s3) [below=of budget5-s3] {\tiny$\mathbf{@s}$};
        \node[state] (s4) [below=of budget5-s4] {\scriptsize$\bot$};
        \node[state,fill=red!35] (s5) [below=of budget5-s5] {\tiny$\mathbf{@s}$};
        \node[state,fill=red!35] (s6) [below=of budget5-s6] {\tiny$\mathbf{@s}$};
        \node[state] (s7) [below=of budget5-s7] {\obsGoalSS};
        \node[state] (s8) [below=of budget5-s8] {\scriptsize$\bot$};
        \node[state] (s9) [below=of budget5-s9] {\scriptsize$\bot$};
        \node[state] (s10) [below=of budget5-s10] {\scriptsize$\bot$};
        \node[state] (s11) [below=of budget5-s11] {\scriptsize$\bot$};
        \node[state] (s12) [below=of budget5-s12] {\scriptsize$\bot$};
        \node[state] (s13) [below=of budget5-s13] {\scriptsize$\bot$};
        \node[state] (s14) [below=of budget5-s14] {\scriptsize$\bot$};
    \end{scope}

    \begin{scope} [name prefix=budget3-]
        \node[state] (s0) [below=of budget4-s0] {\scriptsize$\bot$};
        \node[state] (s1) [below=of budget4-s1] {\scriptsize$\bot$};
        \node[state,fill=red!35] (s2) [below=of budget4-s2] {\tiny$\mathbf{@s}$};
        \node[state] (s3) [below=of budget4-s3] {\scriptsize$\bot$};
        \node[state,fill=red!35] (s4) [below=of budget4-s4] {\tiny$\mathbf{@s}$};
        \node[state] (s5) [below=of budget4-s5] {\scriptsize$\bot$};
        \node[state,fill=red!35] (s6) [below=of budget4-s6] {\tiny$\mathbf{@s}$};
        \node[state] (s7) [below=of budget4-s7] {\obsGoalSS};
        \node[state] (s8) [below=of budget4-s8] {\scriptsize$\bot$};
        \node[state] (s9) [below=of budget4-s9] {\scriptsize$\bot$};
        \node[state] (s10) [below=of budget4-s10] {\scriptsize$\bot$};
        \node[state] (s11) [below=of budget4-s11] {\scriptsize$\bot$};
        \node[state] (s12) [below=of budget4-s12] {\scriptsize$\bot$};
        \node[state] (s13) [below=of budget4-s13] {\scriptsize$\bot$};
        \node[state] (s14) [below=of budget4-s14] {\scriptsize$\bot$};
    \end{scope}

    \begin{scope} [name prefix=budget2-]
        \node[state] (s0) [below=of budget3-s0] {\scriptsize$\bot$};
        \node[state] (s1) [below=of budget3-s1] {\scriptsize$\bot$};
        \node[state] (s2) [below=of budget3-s2] {\scriptsize$\bot$};
        \node[state,fill=red!35] (s3) [below=of budget3-s3] {\tiny$\mathbf{@s}$};
        \node[state] (s4) [below=of budget3-s4] {\scriptsize$\bot$};
        \node[state,fill=red!35] (s5) [below=of budget3-s5] {\tiny$\mathbf{@s}$};
        \node[state] (s6) [below=of budget3-s6] {\scriptsize$\bot$};
        \node[state] (s7) [below=of budget3-s7] {\obsGoalSS};
        \node[state] (s8) [below=of budget3-s8] {\scriptsize$\bot$};
        \node[state] (s9) [below=of budget3-s9] {\scriptsize$\bot$};
        \node[state] (s10) [below=of budget3-s10] {\scriptsize$\bot$};
        \node[state] (s11) [below=of budget3-s11] {\scriptsize$\bot$};
        \node[state] (s12) [below=of budget3-s12] {\scriptsize$\bot$};
        \node[state] (s13) [below=of budget3-s13] {\scriptsize$\bot$};
        \node[state] (s14) [below=of budget3-s14] {\scriptsize$\bot$};
    \end{scope}

    \begin{scope} [name prefix=budget1-]
        \node[state] (s0) [below=of budget2-s0] {\scriptsize$\bot$};
        \node[state] (s1) [below=of budget2-s1] {\scriptsize$\bot$};
        \node[state] (s2) [below=of budget2-s2] {\scriptsize$\bot$};
        \node[state] (s3) [below=of budget2-s3] {\scriptsize$\bot$};
        \node[state,fill=red!35] (s4) [below=of budget2-s4] {\tiny$\mathbf{@s}$};
        \node[state] (s5) [below=of budget2-s5] {\scriptsize$\bot$};
        \node[state] (s6) [below=of budget2-s6] {\scriptsize$\bot$};
        \node[state] (s7) [below=of budget2-s7] {\obsGoalSS};
        \node[state] (s8) [below=of budget2-s8] {\scriptsize$\bot$};
        \node[state] (s9) [below=of budget2-s9] {\scriptsize$\bot$};
        \node[state] (s10) [below=of budget2-s10] {\scriptsize$\bot$};
        \node[state] (s11) [below=of budget2-s11] {\scriptsize$\bot$};
        \node[state] (s12) [below=of budget2-s12] {\scriptsize$\bot$};
        \node[state] (s13) [below=of budget2-s13] {\scriptsize$\bot$};
        \node[state] (s14) [below=of budget2-s14] {\scriptsize$\bot$};
    \end{scope}

  \foreach \b in {1,...,7} {
  \foreach \i in {0,...,13} {

    \ifnum\i=7\relax\else
      \ifnum\i=6\relax
        \draw[->]
          (budget\b-s\i)
          to node[above] {}
          (budget\b-s\the\numexpr\i+1\relax);
      \else
        \draw[->, bend left=20]
          (budget\b-s\i)
          to node[above] {}
          (budget\b-s\the\numexpr\i+1\relax);
      \fi
    \fi
    
    \ifnum\i=6\relax\else
      \ifnum\i=7\relax
        \draw[->]
          (budget\b-s\the\numexpr\i+1\relax)
          to node[below] {}
          (budget\b-s\i);
      \else
        \draw[->, bend left=20]
          (budget\b-s\the\numexpr\i+1\relax)
          to node[below] {}
          (budget\b-s\i);
      \fi
    \fi
    }
    }
\end{tikzpicture}
\caption{Observation function with lowest minimal expected reward for $\mdpline$ scaled to 15 states for varying budgets.}
\label{fig:optimal-sensor-placement}
\end{figure}

We proceeded to investigate if there are any common patterns for varying $\mdpline$ sizes, budgets, and goal positioning. \Cref{fig:optimal-sensor-placement} illustrates this experiment for $\mdpline$ scaled to 15 states. Every row corresponds to a certain budget $B$ value, starting at $B=7$ and ending at $B=1$. Note that there are 2 symmetric solutions on each side of the goal for every budget, but we choose to illustrate the left one. We also label states with $\mathbf{@s}$ if their sensor is activated, and omit self loops for conciseness. After performing this experiment for \SSP \Line instances of sizes up to 19 states, we observed some common patterns.

First, the best sensor placement is on a single side of the goal even for budgets strictly less than the \textbf{MPB}. Additionally, the sensors are mostly spread around one side increasing coverage, although they are still close to each other. The sensors activated are closer to the goal than to the end of the line, and, finally, there appears to be solution compositionality for budgets $7,\dots,4$. Although some of the patterns identified can be utilized as heuristic components for informed search algorithms, they illustrate the complexity of \OOP variants and the challenges involved in developing efficient algorithms to solve them.

\clearpage
\section{Appendix: Heuristics for Decomposition}\label{app:heuristics-decomposition}

\subsection{Edge cases for $A_\mathfrak{G}$}\label{app:incompleteness-clustering}

\begin{wrapfigure}{r}{0.26\textwidth}
\vspace{-0.72cm}
\begin{tikzpicture}[->,>=stealth',shorten >=1pt,auto,node distance=1.3cm,semithick]
  \tikzstyle{every state}=[fill=white,draw=black,text=black]

  \node[state,inner sep=2pt,minimum size=1pt] (S0)                   {$s_0$};
  \node[state,inner sep=2pt,minimum size=1pt] (S1) [below of=S0]     {$s_1$};
  \node[state,inner sep=2pt,minimum size=1pt] (TRAP) [below left of=S1] {$s_2$};
  \node[state,inner sep=2pt,minimum size=1pt][fill=green!20] (GOAL) [below right of=S1] {$s_3$};
  \node[state,inner sep=2pt,minimum size=1pt] (S2) [below right of=TRAP]     {$s_4$};
  \node[state,inner sep=2pt,minimum size=1pt] (S3) [below of=S2]     {$s_5$};

  \path (S0) edge [bend left=20]             node[right] {$c$} (GOAL)
             edge  [bend right=20]             node [left]{$b$} (TRAP)
             edge  [loop below]             node {$a$} ()
             
        (S1) edge [bend left=10]             node[above left=1pt and -6pt] {$a$} (GOAL)
             edge  [bend right=10]             node [above left=1pt and -6pt]{$b$} (TRAP)
             edge  [loop below]             node[below right=-3pt and 1pt] {$c$} ()

        (S2) edge [bend right=10]             node[below left=1pt and -6pt] {$a$} (GOAL)
             edge  [bend left=10]             node [below left=1pt and -6pt]{$c$} (TRAP)
             edge  [loop above]             node[above left=-3pt and 1pt] {$b$} ()

        (S3) edge [bend right=20]             node [right]{$b$} (GOAL)
             edge  [bend left=20]             node [left]{$c$} (TRAP)
             edge  [loop above]             node {$a$} ()
        (TRAP) edge  [loop left]             node {} ();
\end{tikzpicture}
\caption{MDP $\mdptrap$.}
\vspace{-5pt}
\label{fig:trap-world}
\end{wrapfigure}

Consider $\mdptrap$ in \Cref{fig:trap-world}, with initial states $s_0, s_1, s_4, s_5$. $s_2$ is a ``trap'' from where an agent cannot escape and $s_3$ is the goal, so $\alpha^{*}(s_0) = \{c\}, \alpha^{*}(s_1) = \alpha^{*}(s_4) = \{ a \}, \alpha^{*}(s_2) = \{ a,b,c \}, \alpha^{*}(s_5) = \{ b \}$ and
\vspace{-0.16cm}
\[
\mathfrak{G}_{\mdptrap} = \left \{\{s_0\}, \{s_1, s_4\}, \{ s_2 \}, \{s_5\} \right \}.
\vspace{-0.16cm}
\]
Since $s_1$ and $s_4$ belong to the same atomic distinguishability group, the algorithm only considers observation functions that assign them the same observation; however, the \POP-instance $(\mdptrap, 2, \tau)$ only has a valid certificate $\obs$ if $\obs(s_1) \neq \obs(s_4)$. For contradiction, suppose $\obs$ assigns $\{s_1,s_4\}$ and $\{s_0\}$ into the same observation class. Then, the states in it must do $c$ with some probability, but that would take an agent from $s_4$ into the trap, resulting in an infinite reward. Symmetrically, if $\obs$ combines $\{s_1,s_4\}$ and $\{s_5\}$ into the same class. The only remaining case without breaking $\obs(s_1)=\obs(s_4)$ is to have $s_0$ and $s_5$ in the same observation class. However, the states in it need to do both $b$ and $c$ with some probability, which again leads an agent into the trap. Therefore, $A_\mathfrak{G}$ never explores the only valid certificate for the \POP-instance $(\mdptrap, 2, \tau)$ and, as such, is not complete. To improve robustness in our implementation, we added a feature to solve for the entire \tpmc encoding of the instance after the main loop of the algorithm. 

\subsection{Further Applications of Heuristics for Decomposition}\label{app:mpbp-initial-guess}

As concluded in \cite{cav-base-oop}, the main idea for solving the problem \textbf{MPBP} is that every optimal, positional, and deterministic (\textsc{opd}, for short) strategy $\sigma\colon S_\mdp \to \Act_\mdp$ for an MDP $\mdp$ also solves \textbf{PDOOP} for $\mdp$ with threshold $\tau = \MinExpRewP{\mdp}$ and budget $B = |\mathit{range}(\sigma)|$:
A suitable observation function $\obs\colon S_\mdp \to \mathit{range}(\sigma)$ assigns action $\alpha$ to every state $s \in S_\mdp$ with $\sigma(s) = \alpha$.
It thus suffices to find an \textsc{opd} strategy for $\mdp$ that uses a minimal set of actions.
A brute-force approach to finding such a strategy iterates over all subsets of actions $A \subseteq \Act_\mdp$: For each $A$, we construct an MDP $\mdp_A$ from $\mdp$ that keeps only the actions in $A$, and determine an \textsc{opd} strategy $\sigma_A$ for $\mdp_A$. 
The desired strategy is then given by the strategy for the smallest set $A$ such that $\ExpRew{\mdp_A}{\sigma_A} = \MinExpRewP{\mdp}$.
Since finding an \textsc{opd} strategy for a MDP is possible in polynomial time (cf. \cite{baier2008principles}), the problem \textbf{MPBP} can be solved in $O(2^{|\Act_\mdp|} \cdot \textit{poly(size(\mdp)))}$.

\subsubsection{Alternative Algorithm for the MPBP (cf. \Cref{def:minimalBudget})}
Inspired by the algorithm $A_\mathfrak{G}$ introduced in \Cref{subsec:informed-search-pop}, we derive an alternative procedure to solving the \textbf{MPBP} for a \POP-instance $(\mdp, B, \tau)$.

Perform the same ranking of observation functions as $A_\mathfrak{G}$ for $(\mdp, b, \tau),~ b = 1, \dots, k$ where $k = \min \left ( \lvert \Act_\mdp \rvert, \lvert \mathfrak{G}(M) \rvert \right )$, and return the lowest $b$ for which there exists an optimal observation function $\obs\colon S_\mdp \to \OO, \lvert O \rvert = b$. As shown in \Cref{subsec:informed-search-pop}, $\obs$ is optimal iff $\mathit{eq}(\obs) = b$. For improved performance, a binary search can trivially be applied. Simply put, the procedure uses the strong heuristics in $A_\mathfrak{G}$ to solve the \textbf{MPBP} and is most efficient when the number of atomic distinguishability groups is small, such as the case of the \Line, \Maze, and \Grid worlds studied.

\subsubsection{Initial Guesses}
We focus now on creating meaningful initial guesses for any algorithm that explores the state space of possible observation functions for any \POP- or \SSP-instance $(\mdp, B, \tau)$. For \POP, a promising initial guess is the first-ranked observation function evaluated in $A_\mathfrak{G}$. For the latter, we briefly describe below a simple initialization approach based on heuristics collected from our experiments.

First, compute the \textbf{MPB} for the corresponding \POP instance $(\mdp, B, \tau)$ using the most promising algorithm for \textbf{MPBP}, and select the first-ranked observation function $\obs_{\text{POP}}\colon S_\mdp \to O \uplus \{ \obsGoal \}$, where $O$ is a finite set of \textit{observations}. We must then transform $\obs_{\text{POP}}$ into a valid \SSP observation function $\obs_{\text{SSP}}\colon S_\mdp \to D \uplus \{ \bot \}$, where $D \subseteq \{ @s \mid s \in (S_\mdp \setminus G_\mdp) \}$ are the \textit{observable locations}. To do so, we activate the sensor ($\obs_{\text{SSP}}(s) = @s$) for states assigned to the smallest observation classes in $\obs_{\text{POP}}$. More specifically, let us define the size of an observation class $o$ to be $\lvert \obs^{-1}_{\text{POP}}(o) \rvert$ and denote the number of states with a sensor activated as $S_\text{on}$. 
Then, sort $o \in O$ by observation class size. For each observation class $o$ (in ascending order), activate sensors for all states in $o$ if they can be accommodated in the currently available budget $B - S_\text{on} > \lvert \obs^{-1}_{\POP}(o)\rvert$, otherwise uniformly at random states whose activated sensors should fulfill the budget while $S_\text{on} < B$.

Once $S_\text{on} = B$, set $\obs_{\text{SSP}}(s) = \bot$ for the remaining states in the largest observation classes. As a result, $\obs_{\text{SSP}}$ has an assignment for every state in $S_\mdp$ and $\lvert D \rvert = B$, which forms a valid location POMDP $\mdp\setObs{\obs_{\text{SSP}}}$ that serves as an initial guess.

The intuition behind this approach is to organize the largest atomic distinguishability groups for the \POP-instance in the same ``unknown'' class ($\bot$) in the \SSP-instance's observation function, thereby attempting to reduce the number of non-optimal actions that must be performed for states with sensor off. Further domain-specific initialization heuristics can be applied which leverage the topology of the worlds. For instance, in the \Line, \Maze, and \Grid words, the underlying and implicit notion of cardinal directions in the action set implies that actions can be in strict opposite directions from each other (such as $\mathit{up}$ and $\mathit{down}$). One example of additional heuristics is to prioritize activating sensors for states that have an optimal signature that is ``opposite'' to the set of local optimal actions for states assigned to $\bot$.

\clearpage

\end{document}